\documentclass[runningheads]{llncs}

\usepackage{array}
\usepackage{multirow}
\usepackage{textcomp}
\usepackage{booktabs}
\usepackage{graphicx}
\usepackage{tabularx}
\usepackage{pdfpages}
\usepackage{tikz}
\usepackage{tablefootnote}
\usepackage{textcomp}
\usepackage{verbatim}
\usepackage{makecell}
\usepackage{enumitem}
\usepackage{comment}
\usepackage{subcaption}
\usepackage[T1]{fontenc}
\usepackage{lscape}
\usepackage[list=true]{subcaption}
\usepackage{hyperref}
\usepackage{textcomp}

\begin{document}
\setcounter{secnumdepth}{3}

\pagestyle{headings}
\mainmatter
\def\ECCVSubNumber{100}  


\title{Deep Learning Approaches for \\ Seizure Video Analysis: A Review}

\titlerunning{Deep Learning Approaches for Seizure Video Analysis: A Review}

\author{David Ahmedt-Aristizabal\inst{1,2} \and
Mohammad Ali Armin\inst{1} \and
Zeeshan Hayder\inst{1} \and
Norberto Garcia-Cairasco\inst{3} \and
Lars Petersson\inst{1} \and
Clinton Fookes\inst{2} \and
\\ Simon Denman\inst{2} \and
Aileen McGonigal\inst{4,5}
}
\authorrunning{Ahmedt-Aristizabal et al.}

\institute{Imaging and Computer Vision Group, CSIRO Data61, Australia 
\email{\{david.ahmedtaristizabal; ali.armin; zeeshan.hayder; lars.petersson\}@csiro.au}
\and
SAIVT Laboratory, Queensland University of Technology, Australia\\
\email{\{c.fookes; s.denman\}@qut.edu.au} 
\and
Physiology Department and Neuroscience and Behavioral Sciences Department, Ribeirão Preto Medical School, University of São Paulo, Brazil \\
\email{ngcairas@usp.br} 
\and
Neurosciences Centre, Mater Hospital, Australia
\and
Queensland Brain Institute, The University of Queensland, Australia\\
\email{a.mcgonigal@uq.edu.au} 
}

\maketitle




\vspace{-12pt}
\begin{abstract}
\scriptsize Seizure events can manifest as transient disruptions in the control of movements which may be organized in distinct behavioral sequences, accompanied or not by other observable features such as altered facial expressions. The analysis of these clinical signs, referred to as semiology, is subject to observer variations when specialists evaluate video-recorded events in the clinical setting. To enhance the accuracy and consistency of evaluations, computer-aided video analysis of seizures has emerged as a natural avenue.
In the field of medical applications, deep learning and computer vision approaches have driven substantial advancements. Historically, these approaches have been used for disease detection, classification, and prediction using diagnostic data; however, there has been limited exploration of their application in evaluating video-based motion detection in the clinical epileptology setting.
While vision-based technologies do not aim to replace clinical expertise, they can significantly contribute to medical decision-making and patient care by providing quantitative evidence and decision support. Behavior monitoring tools offer several advantages such as providing objective information, detecting challenging-to-observe events, reducing documentation efforts, and extending assessment capabilities to areas with limited expertise. The main applications of these could be (1) improved seizure detection methods; (2) refined semiology analysis for predicting seizure type and cerebral localization.
In this paper, we detail the foundation technologies used in vision-based systems in the analysis of seizure videos, highlighting their success in semiology detection and analysis, focusing on work published in the last 7 years. We systematically present these methods and indicate how the adoption of deep learning for the analysis of video recordings of seizures could be approached. Additionally, we illustrate how existing technologies can be interconnected through an integrated system for video-based semiology analysis. Each module can be customized and improved by adapting more accurate and robust deep learning approaches as these evolve. Finally, we discuss challenges and research directions for future studies. 
%

\vspace{-5pt}
\end{abstract}




\section{Introduction}
Seizure semiology involves the analysis of seizure clinical manifestations, which may include stereotypical behavioral and motor changes, among them tonic, clonic or complex motor patterns. 
The clinical expression of seizure represents the primary symptomatic challenge in epilepsy. Various semiologic manifestations of seizures are potentially measurable using video analysis: motor alterations (\textit{e.g.}~as seen in automatic behaviors, hyperkinetic seizures, or tonic-clonic seizures); an abnormal lack of movement (typical of some seizures with prominent loss of responsiveness to the environment); or an abrupt change in posture that constitutes a risk factor for falls (\textit{e.g.}~in atonic seizures).
Alongside electrophysiological recordings and neuroimaging, the insights derived from semiology are essential, particularly for patients resistant to drug treatment~\cite{bonini2014frontal,thijs2019Epilepsy,alim2022probabilistic}. 
Seizure semiology, though a crucial data source for clinicians, poses substantial methodological challenges for research, such as the selection of semiologic categories. A more profound understanding of semiologic categorization holds relevance for refining seizure classification systems~\cite{luders1999new,mcgonigal2021seizure}. 
The interpretation of video monitoring recordings, however, relies on clinical experience and can vary between clinicians and across patients, especially when seizures involve complex signs including hyperkinetic motor behavior~\cite{seneviratne2012good}.

Therefore automated quantification of semiology shows promise for extracting more objective information from these recordings~\cite{ahmedt2019understanding,karacsony2022novel,hou2022automated}. 
Quantified analysis of seizures through video is desirable for its non-invasiveness and potentially easy deployment, utilizing existing video hardware in various settings, once clinically relevant algorithms have been developed. A main motivation in developing this field has been the goal of accurate automated seizure detection, because of the potential impact on morbidity, mortality and quality of life in people with epilepsy~\cite{knight2023artificial}.

Expanding on the interdisciplinary nature of this research, it is noteworthy that although the characterization of behavior, in general, needs to be placed in the context of biology and specifically of ethology, the systematic study of behavior~\cite{tinbergen2012social} has evolved into quantitative methods such as neuroethology~\cite{fentress1973grammar} and computational neuroethology~\cite{datta2019computational}. These concepts, associated with the study of the complexity of behavioral sequences such as those present in human epilepsies~\cite{garcia1996neuroethological,dal2006neuroethology,bertti2010neurobiological,tejada2013epilepsies}, offer a unique perspective. The application of these approaches to epilepsy research allows for a more integrated understanding of the complex interactions between neural circuits and behavior during seizure activity. Moreover, they provide a solid foundation for developing computational models that bridge the gap between clinical observation and automated quantification.

Automated methods for seizure detection and classification may also help improve patient monitoring, reducing the time and effort needed for screening long-term video-EEG data in specialized Epilepsy monitoring units (EMUs). Better methods for automated video-based seizure detection can also be applied to home recording systems, once validated in a hospital setting~\cite{peltola2022semiautomated}. Lastly, as in other clinical fields, artificial-intelligence-enhanced diagnostic methods may be trained to ``see'' patterns that clinicians do not reliably recognize, with the potential for better prediction of cerebral localization from complex semiologic features, such as those seen in hyperkinetic seizures. This last application in particular is under-developed due to challenges such as the scarcity of datasets and the complexity of the natural clinical setting, which need to be addressed in this field. 

\vspace{-6pt}
\subsection{What is modern deep learning and computer vision?}
The growing use of artificial intelligence (AI) and machine learning has significantly influenced various technological domains, such as signal and human motion analysis. Machine learning, a subset of AI, encompasses diverse methods that enable systems to automatically learn patterns and make predictions from data. The application of these learning methods, particularly in the domain of video analysis, has gained prominence with the advent of modern deep learning approaches. Deep learning, a subset of machine learning, employs many-layered neural networks to automatically learn hierarchical features from data~\cite{lecun2015deep}. Notably, the rise of deep learning has seen the field of video-based human action analysis become a focal point within the field of computer vision and pattern recognition, with recent technical advances allowing increasingly detailed analyses of actions, gestures, and facial expressions~\cite{wu2017recent}.

A foundational learning paradigm in machine learning is supervised learning, where an algorithm is trained on a labeled dataset, with input data and the corresponding desired outputs provided. From this, a model is trained to learn the mapping between inputs and outputs. As such techniques are data-driven and dependent on the quantity and quality of the provided data, there are several critical challenges for medical applications including scarce annotation (too little labelled data), complexity (a single type of instance to detect may appear in many different forms), weak annotations (labels may be noise), and label sparsity (some classes have very few instances).
To address these challenges, several other learning paradigms have been proposed in the literature.
Weakly supervised learning emerges as a strategy, leveraging partially annotated examples to navigate the scarcity of labeled data. Self-supervised learning adopts a different approach, where the data itself provides supervisory signals, aiding in learning representations through proxy tasks. Additionally, reinforcement learning introduces a concept where an agent learns optimal behavior by interacting within a dynamic environment.
Further background concepts in the domain of computer vision and deep learning are discussed in Section~\ref{background}.

\vspace{-6pt}
\subsection{Vision-based action recognition in health}
Human action recognition, a key aspect of semiology analysis, aims to identify different actions from sequences of observations across diverse environmental conditions. 
The applications of such recognition methods driven by deep learning and computer vision are wide-ranging, contributing to advancements relating to human health and performance. 
Applications include assessment of human development, optimization of human performance, neuromuscular rehabilitation, prevention of musculoskeletal injury, gait analysis, and motor assessment~\cite{stenum2021applications,avogaro2023markerless}. Existing literature extensively explores the causal relationship between diseases and abnormal body actions such as atypical head poses and movements, emotions, eye movements, as well as upper and lower limb movements~\cite{dash2022review,turaev2023review}.

\vspace{-6pt}
\subsection{Vision-based seizure semiology analysis}
There is an increasing interest in leveraging video analysis techniques for the automated analysis of movements associated with motor stereotyped manifestations such as those from Parkinson's disease, tremors, peripheral neuropathy, Huntington's chorea, ataxia, dystonia, and epilepsy~\cite{do2016movement,mesquita2019methodological,myszczynska2020applications,javeed2023machine}. Vision-based techniques, in particular, have proven valuable in patient monitoring~\cite{sathyanarayana2018vision}.
While general vision-based motion analyses excel in uncluttered settings, they can be unreliable in noisy environments like epilepsy monitoring and intensive care units. These settings pose unique challenges, including varying lighting conditions throughout recording sessions, environmental occlusions (\textit{e.g.}~bed blankets, head wrapping, hospital gown), and interference from non-subject entities (\textit{e.g.}~clinician, nurse, visitor)~\cite{tian2020automated}.
While current state-of-the-art deep learning models exhibit promise in addressing these challenges in epilepsy, their effectiveness is still in the early stages. We can anticipate that more advanced video-based automatic semiologic categorization will become possible as the field progresses. 
Existing methods struggle to recognize subject-specific semiologic categories of unseen behaviors, yielding modest performance. Achieving fine-grained semiology recognition, crucial for distinguishing the stepwise progression of clinical features, remains a challenge. In addition, ensuring effective explainability to understand model behavior beyond traditional performance indicators is essential, with a particular focus on clinician usability. As these translations of research progress, there is a need for research to facilitate the proper clinical adoption of methods and to build trust between clinical experts and AI frameworks.

\vspace{-6pt}
\subsection{Scope of review and systematic review methodology}
As the field is rapidly evolving and researchers are interested in understanding the strengths and pitfalls of deep learning-based solutions for the analysis of seizure semiology, a comprehensive survey is required that provides insights from existing studies and motivates researchers to address limitations and challenges within this field. 
Conventional motion analysis of seizures, employing classical markerless computer vision methods, was reviewed by Pediaditis et al.~\cite{pediaditis2010vision} and Abbasi et al.~\cite{abbasi2019machine}. 
A more detailed examination, considering data-driven solutions for motion capture and action detection, was presented by Ahmedt-Aristizabal et al.~\cite{ahmedt2017automated}. 
Karacsony et al.~\cite{karacsony2023deep} provided insights into advancements in video-based clinical in-bed monitoring, particularly within EMUs. 
These works discuss opportunities to address challenges associated with 3D motion capture, including domain transfer, occlusions, varying viewpoints, and dealing with low resolution footage. A recent study~\cite{knight2023artificial} outlined the anticipated workflow of a ``complete system'' for video analysis of seizure event classification. However, this non-technical discussion overlooked crucial stages. 

The application of deep learning approaches to seizure video analysis is still in its nascent stages. 
The goal of this review is to provide an overview of advances in the field, focusing on work published from 2017 and onwards (the date of our previous review), and highlight ongoing challenges and future directions.
The following search string was used in the Google Scholar search engine as an initial search action:
(human OR patient) AND (Epilepsy OR epileptic) AND (video OR vision OR computer vision OR action recognition) AND (motion OR movement OR seizure OR semiology OR motor) AND (analysis OR detection OR recognition OR classification) AND (automatic OR machine learning OR deep learning OR artificial intelligence). 
Case reports and other deep learning applications in Epilepsy including neuroimaging (\textit{e.g.}~MRI, fMRI) and brain activity (\textit{e.g.}~EEG, iEEG) were excluded. Further papers were identified by searching publications by the same authors of initially identified relevant papers, following references, and using the ``Cited'' function. The results were filtered based on their title and abstract. Case studies reported in this paper are obtained from various peer-reviewed journals, conference proceedings, and open-access repositories published in English. The total number of applications considered in our survey is 35 manuscripts. 

\vspace{-6pt}
\subsection{Contribution and organization}
In contrast to other reviews that focus on theoretical aspects and applications of action recognition in semiology, our manuscript makes novel contributions, summarized as follows:
\vspace{-4pt}
\begin{itemize}
    \item We extend our 2016 review of vision-based approaches for semiology analysis~\cite{ahmedt2017automated} by incorporating all relevant studies concerning the adoption of deep learning techniques to enhance the automated detection and quantification of seizures, and/or of individual seizure features captured by video recordings. All deep learning methods adopted by these works are described in detail and categorized according to the motion analysis framework used.
    \vspace{-2pt}
    \item We introduce a multi-stream framework that connects different advancements in automated semiology analysis, aiming to facilitate the adoption of these technologies by clinical experts to provide diagnostic support.
    \vspace{-2pt}
    \item Our paper identifies several challenges specific to action recognition models in the context of seizure semiology based on video analysis, which have not been comprehensively addressed in previous reviews. These challenges cover areas such as motion capture of stereotyped movements and domain adaptation, privacy preservation, interpretation of action models, open-set action analysis, and time-evolving or fine-grained action recognition. Existing research efforts in AI and computer vision addressing these challenges are highlighted.
\end{itemize}

This review is divided into five major sections.
Section~\ref{neuroethology} elucidates concepts and methods in quantifying behavior and behavioral sequences from a neurobiological/neuroscience perspective. Then, the section explores the application of computational neuroethology tools in evaluating these behavioral patterns.
In Section~\ref{background}, we provide a technical overview of human action recognition (HAR) and current approaches for vision-based seizure semiology analysis. Specifically, Section~\ref{background:pose}, introduces skeleton-based HAR approaches, while Section~\ref{background:spatiotempo} focuses on frame-based HAR models.
In Section~\ref{litreview}, we introduce the current applications of computer vision and deep learning techniques for semiology analysis. Details regarding the existing approaches along with the computer vision techniques are categorized according to the motion framework employed.
Section~\ref{guideline} offers guidance on incorporating existing approaches along with the computer vision techniques they employ for clinical use through a multi-stream framework for semiology analysis. 
Finally, in Section~\ref{opport}, we clarify limitations and offer new perspectives on the current state of semiology-based seizure analysis.


\section{Human movement analysis in the epilepsies: from traditional clinical observation to quantitative and computational neuroethology}
\label{neuroethology}

The evolution of technology has played a pivotal role in shaping our understanding of human behavior, especially in the context of conditions such as motor epilepsies. Computational tools have become integral in uncovering the origins, triggers, and underlying mechanisms of epileptic manifestations, marking a significant advancement in neuroscience through digital and computational technology.

In a recent review~\cite{garcia2021searching}, we proposed transitioning from traditional clinical observations to quantitative semiology, acknowledging the challenge of unknown neural substrates in earlier eras.
This paradigm shift aligns with the systematic evaluation of behavior in animals and humans within the science of ethology~\cite{tinbergen2012social}, involving the creation of dictionaries for detailed behavior descriptions.
In the particular case of characterizing semiology, focusing on behavioral features, pioneering studies discussed in~\cite{garcia2021searching} have found detailed associations, distinguishing patients with epilepsy from similar clinical entities such as psychogenic non-epileptic seizures (PNES) or cases with neuropsychiatric dysfunctions.

In the progression of this research, the concept of neuroethology emerges as the search for neural substrates of behaviors~\cite{fentress1973grammar}. Expressions such as the ``Grammar of a movement sequence'' have been utilized by authors in the field to describe this pursuit. 
Laboratories initiated the development of ethograms and behavioral glossaries~\cite{garcia1983role}. These protocols involved annotating every behavior and its probabilistic interactions displayed by animals when stimulated, modeling various seizure types~\cite{garcia2017wistar,furtado2011study,castro2011comparative}. While the term was not explicitly used then, these practices formed the foundation for what we now refer to as quantitative human neuroethology.

Quantitative neuroethology of behavioral sequences served as the natural tool for studies, particularly in epileptic seizures. Initially conducted using animal models~\cite{garcia1992new,garcia1996neuroethological}, these studies were later extended to human patients with both temporal lobe~\cite{dal2006neuroethology,bertti2010neurobiological} and frontal lobe~\cite{bertti2014looking} epileptic seizures.
For example, behavioral sequences were reconstructed as flow charts or graphs, with measurements of complexity associated with the expression of the respective semiology~\cite{tejada2013epilepsies}. 
Other researchers have proposed a paradigm shift, viewing semiology as being dynamically produced by interconnected structures, emphasizing the role of specific rhythmic interactions between connected structures within specific networks in clinical expression~\cite{mcgonigal2021seizure}. 

An ongoing challenge is the expression of overlapping/redundant substrates of associated interictal neuropsychiatric comorbidities in the behavioral aspects of epilepsy.
For example, studies with the concept of autobiographical memories~\cite{rayner2017contribution} are vital, yet are often excluded in the criteria used to define epileptogenic regions.

Building upon these foundations, we believe that evaluating complex experimental/human behavioral sequences in the context of epilepsy semiology can be significantly enhanced with the deep learning and computer vision approaches discussed in this review. 
Strengthening this proposal, Krakauer et al.~\cite{krakauer2017neuroscience} emphasized the importance of behavior studies in neuroscience, advocating for a departure from contemporary reductionistic molecular/cellular biases.

Another challenge and intriguing pathway for advancements in human studies is the adaptation of the concept of quantitative neuroethology to computational neuroscience and modeling. Coined as computational neuroethology~\cite{datta2019computational,wiltschko2015mapping}, this concept focuses on automatically reconstructing complex behavioral fragments, behaviors, or sequences through digital video recording and analysis of individuals or groups of experimental animals. 
A pioneering program in this field, known as MoSeq2~\cite{weinreb2023keypoint}, autonomously collects serial/sequential sets of behaviors. MoSeq2 utilizes a machine learning-based platform to identify behavioral modules (``syllables'') from keypoint data without human supervision. 

In conclusion, the fusion of quantitative neuroethology, computational neuroscience, and advanced technologies, including computer vision and deep learning, emerges as a powerful approach to unravel the intricate semiology of epilepsies. This integrated approach not only contributes significantly to diagnosis and treatment strategies, but also signifies a crucial step forward in addressing the challenges posed by this complex neurological disorder. 



\section{Computer vision for human motion analysis: background}
\label{background}

\subsection{Overview of modern deep learning and human action recognition}
\label{background:terminology}

Deep learning, a subset of machine learning, enables the creation of computational architectures comprising multiple processing layers. These architectures learn data representations with varying levels of abstraction through non-linear transformations~\cite{bengio2013representation}.
The primary distinction from traditional machine learning lies in feature engineering. In deep learning algorithms, feature representations are automatically derived from the training data rather than relying on human assumptions as seen in traditional hand-crafted approaches~\cite{lecun2015deep}. This proves particularly beneficial in scenarios where understanding the relationship between input and output is limited~\cite{arel2010deep}. Deep learning techniques have become predominant in computer vision, outperforming traditional approaches across various tasks, including human action recognition. Noteworthy deep neural network architectures include convolutional neural networks (CNNs), recurrent neural networks (RNNs), graph neural networks (GNNs) and the Transformer model.

Within the broader area of computer vision, Human Action Recognition (HAR) has become a compelling research field. The challenges in action recognition extend beyond the complexities of defining body part motions, encompassing various real-world issues like camera motion and dynamic background fluctuations, as well as variations between subjects. HAR is designed to automatically detect and comprehend actions executed by a human subject, leveraging information captured through a camera. Video-based approaches not only analyze spatial patterns within video frames/images but also learn temporal patterns. Visible (RGB) data is the predominant data type for HAR, widely employed in surveillance and monitoring systems. Human motion analysis, a broader domain, spans applications from fundamental motion detection to intricate tasks such as action recognition, classification, and prediction. For an in-depth exploration of this field, readers are encouraged to refer to survey manuscripts~\cite{jegham2020vision,beddiar2020vision,pareek2021survey,sun2022huamn}.

Deep learning methods may also be trained in different ways. 
In a supervised learning paradigm, training occurs offline as the machine learning processes use the labeled data to learn a mapping from the input to the output, and thus predict output for unforeseen data. However, the scarcity of labeled clinical data and challenges in acquiring additional data make efficient data utilization crucial. To overcome the limitations posed by small datasets, HAR architectures rely on transfer learning. 
Networks are pre-trained on large-scale datasets~\cite{ghadiyaram2019large} which allows the architecture to learn a robust general representation, which can then be applied to extract features and learn a representation specifically for the target domain such as seizure video recordings. However, the generalizability of trained classifiers in subject-independent classification settings is hindered by the considerable variation in behavior across different subjects. 
A research direction with significant potential is self-supervised learning. In scenarios where no class labels are available, an effective representation could potentially be learned through an end-to-end method entirely in an unsupervised manner, utilizing a large volume of unlabeled data.

\vspace{-6pt}
\subsection{Overview of vision-based seizure semiology analysis}
\label{background:overview}

Seizures characterized by motor manifestations are analyzed and can be automatically classified based on the specific type of visible (mainly motor) symptomatology. 
Some seizures are defined by unnatural elementary movements, such as those seen in myoclonic, clonic, tonic, versive, and tonic-clonic seizures~\cite{fisher2017operational}. 
The machine-based classification depends on factors such as the duration of muscle contraction, rhythmicity of movement repetition, and the muscle groups or body segments involved (\textit{e.g.}~asymmetric posture, flexion of the neck, abduction of both arms, and turning of eyes and head to one side). 
In contrast, more complex motor manifestations may sometimes seem naturalistic, and usually comprise action patterns involving movements with more complex temporal organization involving different body segments (\textit{e.g.}~manual and oral automatisms such as chewing, swallowing, smacking the lips and fumbling with clothes)~\cite{noachtar2009semiology,blume2001glossary}.

Some existing research on automated quantification of semiology is based on the hypothesis that seizures with similar motor symptoms may involve neuronal activity within the same specific brain networks, and thus could be sufficient to detect and categorize patients with specific types and brain localizations of epilepsy~\cite{knight2023artificial}. 
Vision-based approaches for semiology highlight the significance of analyzing spatio-temporal information during seizures (\textit{i.e.}~behavioral sequences), employing action recognition methodologies. In this domain, the discriminative power of deep learning architectures is exploited for markerless motion capture (MoCap) of body regions for isolated clinical manifestations from the body, head, face, and hands. Furthermore, deep learning is integral to seizure detection (identifying epileptic events within a continuous data stream) and seizure classification (categorizing seizures or seizure disorders). 

Deep learning models play a crucial role in advancing motion capture technologies, particularly for detecting regions of interest (ROI). These ROIs may encapsulate entities like patients, heads, faces, or hands. Yet these techniques go beyond mere object identification and localization, inferring facial landmarks, body keypoints, and hand landmarks from video data. 
While seizures, strictly speaking, may not be categorized as actions, HAR models are designed to encode robust representations of human motion that may prove relevant for seizure characterization. These methods are commonly structured to classify human actions by aggregating predictions for snippets sampled from short clips. Traditionally, a subject-level prediction is obtained by averaging all predictions made at the snippet level.

\begin{figure*}[!t]
\centering
\includegraphics[width=0.98\linewidth]{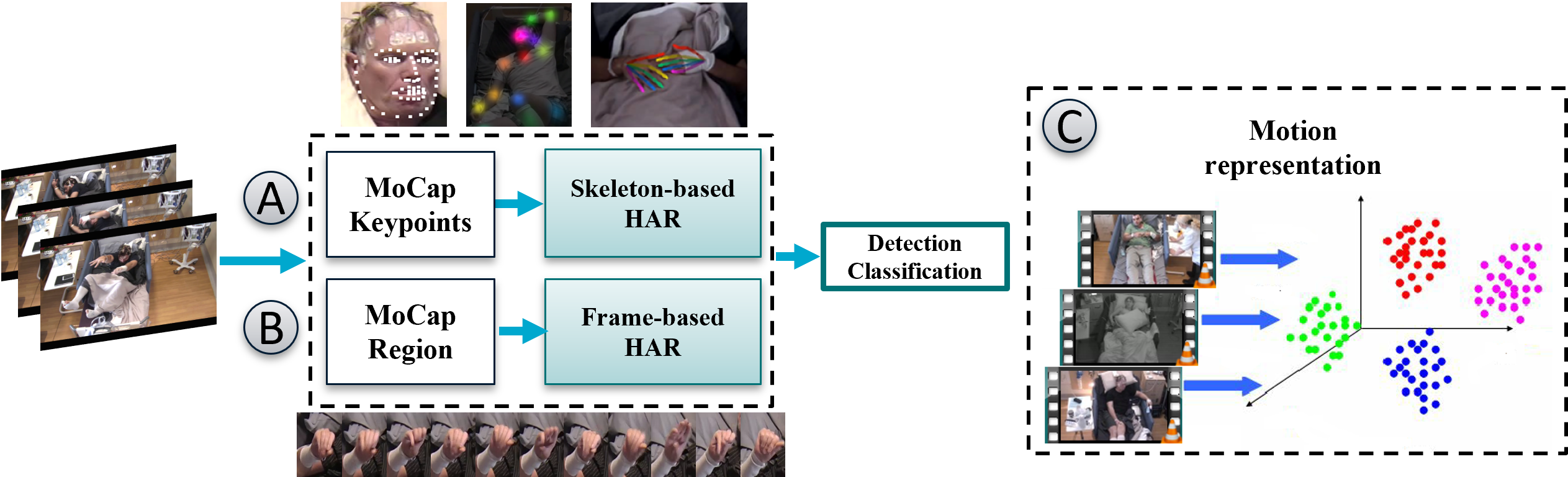}
\vspace{-3pt}
   \caption{Overview of deep learning applications for semiology analysis.
   \textbf{A.} Motion capture and skeleton-based human action recognition.
   \textbf{B.} Frame-based human action recognition.
   \textbf{C.} Feature representations of the motion exhibited by a patient during a seizure, which is used to identify if new data samples are similar to known semiologies, enabling the categorization of Epilepsy types.
   }
\label{fig:AI_semiology}
\vspace{-8pt}
\end{figure*}

In the pre-deep learning era, various hand-crafted feature-based approaches, such as space-time volume-based methods, space-time interest point-based methods, and trajectory-based methods, were employed for RGB video-based detection and HAR. However, with the advent of deep learning, contemporary research in semiology has shifted to the adoption of different types of deep learning frameworks. 
These frameworks fall into two categories, skeleton-based and frame-based HAR. A visual representation of these frameworks is provided in Figure~\ref{fig:AI_semiology}.
The first, skeleton-based HAR, focuses on detecting semantic key points or movement direction/intensity initially. Subsequently, quantitative kinematic and spectral parameters are computed to detect/classify Epilepsy types with classical machine learning or deep learning methods. Here, skeleton information refers to body, facial, or hand landmarks. These approaches have been utilized by~\cite{ahmedt2018deep,ahmedt2019vision,hypponen2020automatic} (See Figure~\ref{fig:AI_semiology}A).
The second framework, a frame-based HAR, captures the dynamic variation of the body’s physical appearance from sequences of images. This approach considers both the presence of specific movements of interest (MOI) and their dynamics and biomechanical characteristics (\textit{e.g.}~speed/acceleration patterns, movement amplitude).
Works including ~\cite{ahmedt2018deep_face,karacsony2020deep,perez2021transfer,yang2021video,hou2022self} are example of this approach (See Figure~\ref{fig:AI_semiology}B).
Seizure semiology encompasses the study of the stepwise/temporal progression of a constellation of signs that are reflective of dynamic changes within connected neuronal networks. Thus, investigating a single clinical sign in isolation is generally less informative than considering all signs jointly and in the context of their usual temporal relationship. As such, multi-stream frameworks have been proposed, capable of jointly quantifying semiologic features from different body locations through fusion or hierarchical approaches (\textit{e.g.}~an ensemble model learning from pose and face skeleton-based streams).
Hybrid networks, which fuse the benefits of both skeleton-based and frame-based approaches for seizure detection and classification, have also been proposed in~\cite{ahmedt2019understanding,hou2021multi,hou2022automated}.
Characteristic combinations of clinical features of different seizure types can be demonstrated to be associated with seizures arising from different brain regions at group level, although predictive power and degree of specificity/sensitivity of these varies. For some types of epilepsy, it is the motor features of the seizures that seem to most strongly predict brain localization; this has been demonstrated at the sub-lobar level in frontal lobe epilepsy~\cite{bonini2014frontal}. Such seizure types with prominent and potentially discriminating motor patterns lend themselves well to analysis using computer vision-based approaches. A visual example of these motions is illustrated in Figure~\ref{fig:AI_semiology}C. The utilization of these feature representations contributes to the development of robust models capable of discerning and categorizing diverse seizure types.
Further details on recent case studies of vision-based approaches in Epilepsy are discussed in Section~\ref{litreview}. The following subsections provide details of the specific techniques used by each human action recognition framework.

\vspace{-6pt}
\subsection{Skeleton-based human action recognition}
\label{background:pose}

\begin{figure*}[!t]
\centering
\includegraphics[width=0.98\linewidth]{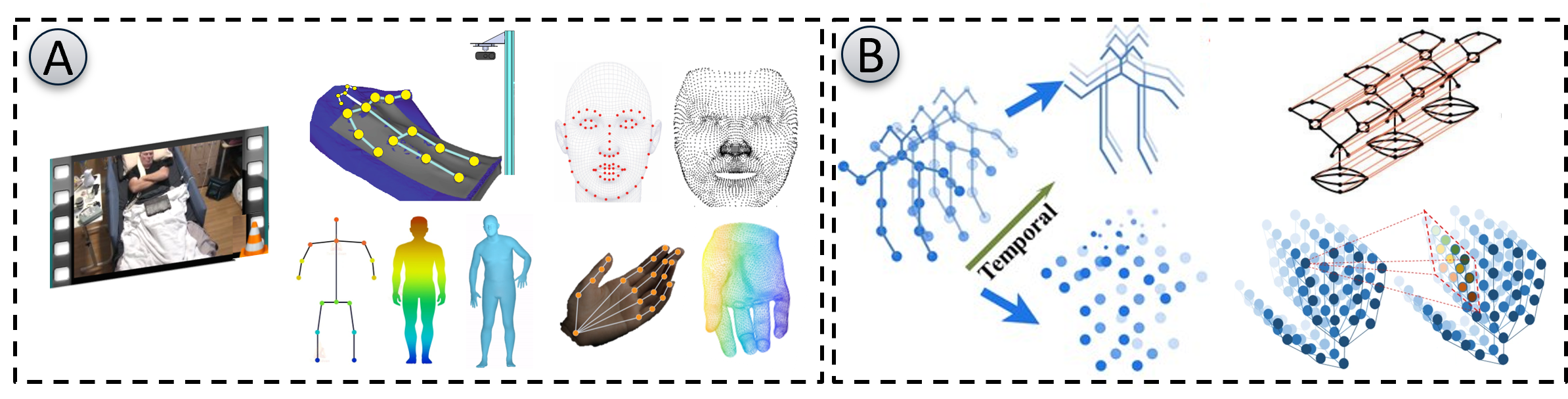}
\vspace{-3pt}
   \caption{Overview of skeleton-based human action recognition.
   \textbf{A.} Motion capture: Includes 2D/3D landmarks and 3D mesh estimation for the body, face and hands. 
   \textbf{B.} Spatio-temporal representation: depicts the motion of each keypoint from the body, face, or hands. RNN-based, CNN-based, GNN-based, and Transformer-based methods are all commonly used with this type of approach.
   }
\label{fig:HAR_approaches1}
\vspace{-8pt}
\end{figure*}

In skeleton-based action recognition, the primary focus is on explicitly tracking and analyzing the movement and positions of skeletal joints from the body, face, or hands, that inherently contain spatio-temporal information. 
Skeleton data offers numerous advantages for HAR including a simple yet informative representation, scale invariance, and robustness to variations in clothing textures and backgrounds. These approaches are often fast since the human pose representation is very compact. However, they lack appearance and detailed shape information and the underlying skeleton representation can be considered a noisy representation~\cite{sun2022huamn}. Another drawback is the propagation of errors from the motion capture to the action recognition stage, a factor highly dependent on the quality of the spatio-temporal feature extraction of the initial stage. Joint information is vulnerable to conditions where joint estimation is performed erroneously, which is not uncommon due to the frequent occlusions that occur in seizure videos. Figure~\ref{fig:HAR_approaches1} illustrates these stages.

\vspace{-6pt}
\subsubsection{Markerless motion capture: keypoint detection}
\label{background:pose_keypoints}

\vspace{4pt}
\noindent \textit{Human pose estimation:}
Human pose estimation involves localizing anatomical keypoints, and is widely studied in computer vision and has broad applications (\textit{e.g.}~motion analysis, healthcare, virtual reality).  Although camera-based methods for human pose estimation have been studied extensively, in-bed pose analysis comes with specific challenges including joints being occluded, blur, and low-light conditions. 
Traditional pose estimation models are broadly categorized into two groups: in the first approach, a human bounding box is initially detected, and then keypoints indicating joint locations for each person are estimated (known as top-down methods, \textit{e.g.}~AlphaPose~\cite{fang2017rmpe}, ViTPose~\cite{xu2022vitpose}). This can lead to errors as the predicted bounding box in the first step may not include the required joints for the second step. In the second approach, all keypoints are detected in the first stage and they are assigned to each person in an image in a second stage (bottom-up methods, or box-free human detection, \textit{e.g.}~OpenPose~\cite{cao2017realtime}).
Human pose estimation has also been investigated in 3D space to predict the locations of body joints. For instance, a 3D human pose estimation algorithm was developed by taking into account spatial and temporal information~\cite{tang20233d}. In addition to estimating the 3D pose, some methods also recover a 3D human mesh from images or videos~\cite{loper2023smpl,pavlakos2019expressive}. This field has attracted much interest in recent years since it is used to provide extensive 3D structure information related to the human body.

\vspace{4pt}
\noindent \textit{Facial landmark estimation:}
Facial information plays a crucial role in semiology analysis. Facial landmark estimators are designed to detect the location of fiducial points, such as the eyes, nose, mouth, or chin, from a face image or video. Facial landmarks can be detected in 2D or 3D space.
2D facial landmark detection and alignment via cascaded regression and CNN-based methods have shown considerable performance~\cite{yan20212d}. However, these methods often regress only the visible points and encounter challenges with large-pose changes and occlusions. To partially address these challenges, facial landmarks can be inferred jointly with their occlusion probability and the overall head pose~\cite{li2023cascaded}. 3D-based methods encompass a wide range of views using a 3D model, which is robust to illumination and pose variations~\cite{bulat2017far,chandran2023continuous}. In addition to these methods, with advancements in neural rendering and generative visual models, synthetic data has gained popularity in facial landmark detection~\cite{zeng20233d}.
The application of more advanced facial landmark methods, especially those demonstrating high performance in handling occlusion and large pose changes, can be a game-changer in extracting richer facial information for semiology analysis. For further information about 2D/3D facial landmarks, consult~\cite{wang2018facial,bodini2019review,sharma20223d,meher2023survey}.

\vspace{4pt}
\noindent \textit{Hand landmark estimation:} 
Hand pose estimation refers to methods used to detect hand landmarks or keypoints. 
This is of interest given the rich and subtle repertoire of abnormal hand postures and movements observed in epileptic seizures, in conjunction with the rapid technical progress in automated methods for hand detection in video~\cite{ferando2019hand,stefan2023ictal}.
Methods that can localize hand joints share several properties with human body pose estimation, and many approaches proposed for the human body can be adapted for hand pose estimation. Similar to body pose, hand pose can also be estimated in 2D~\cite{simon2017hand,wang2018mask} or 3D~\cite{panteleris2018using,cheng2023handr2n2} space. 
Image generation and synthetic data have also become popular in hand pose estimation. Examples of recently introduced synthetic data are ~\cite{khaleghi2022multi}, which includes multi-view hand data, and~\cite{li2023renderih}, which features 1 million photo-realistic images of hands interacting with various objects, backgrounds, and textures.
For more information about 2D/3D or hybrid hand pose models, refer to~\cite{chen2020survey,ohkawa2023efficient}.

\vspace{-6pt}
\subsubsection{Action recognition models (skeleton-based HAR)} 
\label{background:pose_action}

Skeleton-based action recognition centers on extracting meaningful action dynamics from the detected keypoints or dense 3D representations. In the early stages of this field, many attempts encoded all human body joint coordinates in each frame into a feature vector for pattern learning~\cite{wang2019comparative}. Once a representation of the body position has been chosen, various action recognition models come into play, each with its unique approach. These are detailed below.

\vspace{4pt}
\noindent \textit{RNN-based methods:} 
RNN-based methods analyze the vector sequences of joint positions over time. The position of each joint in the human body as it moves over time can be expressed as a vector. The main idea of the RNN is to use a memory-based deep learning module to store information from previous computational steps at time $t-1$ to make the most accurate prediction for the current step at time $t$.
Several studies utilize human pose over time in a recurrent neural network such as Gate Recurrent Unit (GRUs), Long-Short Term Memory (LSTM)~\cite{ye2021deep}, and independently recurrent neural network (IndRNNs)~\cite{li2019deep} and infer the action of interest at every time-step. These approaches are shown to be fast and effective for skeleton-based HAR.

\vspace{4pt}
\noindent \textit{CNN-based methods:} 
This approach utilizes the advantages of CNN-based object recognition in a 2D space (image space). It maps a 3D representation of the 3D human skeletons into a 2D array (possibly capturing the spatial relationships between the skeleton joints) to learn spatio-temporal skeleton features. Examples of recent works using CNN-based spatio-temporal approaches are CNN fusion~\cite{li20213d}, and PoseConv3D~\cite{duan2022revisiting}. 
A CNN fusion model for skeletal action recognition is proposed in~\cite{li20213d}, which uses spatio-temporal information including skeletal trajectory shape images as well as skeletal pose image sequences.  
PoseConv3D~\cite{duan2022revisiting} used the 3D-CNN to capture the spatio-temporal dynamics of skeleton sequences; where the input to the 3D-CNN backbone are 3D heatmap volumes. The pseudo heatmaps for joints and limbs that are generated provide rich information as inputs for 3D-CNNs. More information about convolutional neural network-based action can be found in~\cite{yao2019review}.

\vspace{4pt}
\noindent \textit{GNN-based methods:}
Early approaches to skeleton-based HAR employed RNNs and CNNs, but recent advancements have surpassed them with the introduction of Graph Convolutional Networks (GCNs). Leveraging the expressive power of graph structures, GCNs efficiently represent non-Euclidean data, allowing them to capture both spatial (intra-frame) and temporal (inter-frame) information. These learning models have been adopted for various tasks including skeleton-based human action recognition~\cite{shi2019two,pan2020spatio} and facial landmark-based emotion recognition~\cite{ngoc2020facial}. 
The spatio-temporal graph (ST-GCN)~\cite{yan2018spatial}, the first work utilizing GCNs for HAR, is constructed such that the graph nodes represent the joint and the edges encode natural connections in both human body and time. Multiple layers of spatio-temporal graph convolution operations are applied to the input data, generating higher-level feature maps on the graph, subsequently classifying the input into the corresponding action category. Followed by the success of ST-GCN, actional-structural graphs have been proposed to capture both actional links and structural links over time~\cite{li2019actional}. Another work involves the use of a graph network to perform spatial reasoning and temporal stack learning (SR-TSL) for skeleton-based HAR, comprising a spatial reasoning network (SRN) and a temporal stack learning network (TSLN)~\cite{si2018skeleton}. More recent approaches include ST-GCN++~\cite{duan2022pyskl} and Efficient GCN~\cite{song2022constructing}.

\vspace{4pt}
\noindent \textit{Transformer-based methods:}  
Vision transformers are emerging as a powerful tool to solve computer vision problems. Recent techniques have also proven the efficacy of transformers beyond the image domain, extending to diverse video-related tasks, including action recognition~\cite{arnab2021vivit}.

The architecture of transformers, featuring a self-attention mechanism, proves useful for estimating temporal relationships and motion dynamics encoded in joint positions. Some examples of these approaches include PoseConv3D~\cite{duan2022revisiting}, SkeletonMAE~\cite{wu2023skeletonmae} and InfoGCN++~\cite{chi2023infogcnpp}. PoseConv3D~\cite{duan2022revisiting} addresses limitations in current Skeleton-based Human Activity Recognition (HAR) methods by combining 3D CNN and a transformer-based approach for feature extraction, modeling temporal dependencies, and handling the sparsity of skeleton data.
SkeletonMAE~\cite{wu2023skeletonmae} focuses on capturing structural information and temporal dependencies in skeleton data by representing it as a graph. It introduces masked autoencoders to learn meaningful representations by reconstructing the original skeleton sequences with masked nodes. Furthermore, InfoGCN++~\cite{chi2023infogcnpp}, leverages graph convolutional networks (GCNs) as part of the transformer network to learn informative representations by predicting future frames in the skeleton sequence.
These transformer-based methods incorporate temporal information through a joint learning strategy, resulting in improved performance in action recognition tasks.

\vspace{-6pt}
\subsection{Frame-based HAR modality}
\label{background:spatiotempo}

\begin{figure*}[!t]
\centering
\includegraphics[width=0.85\linewidth]{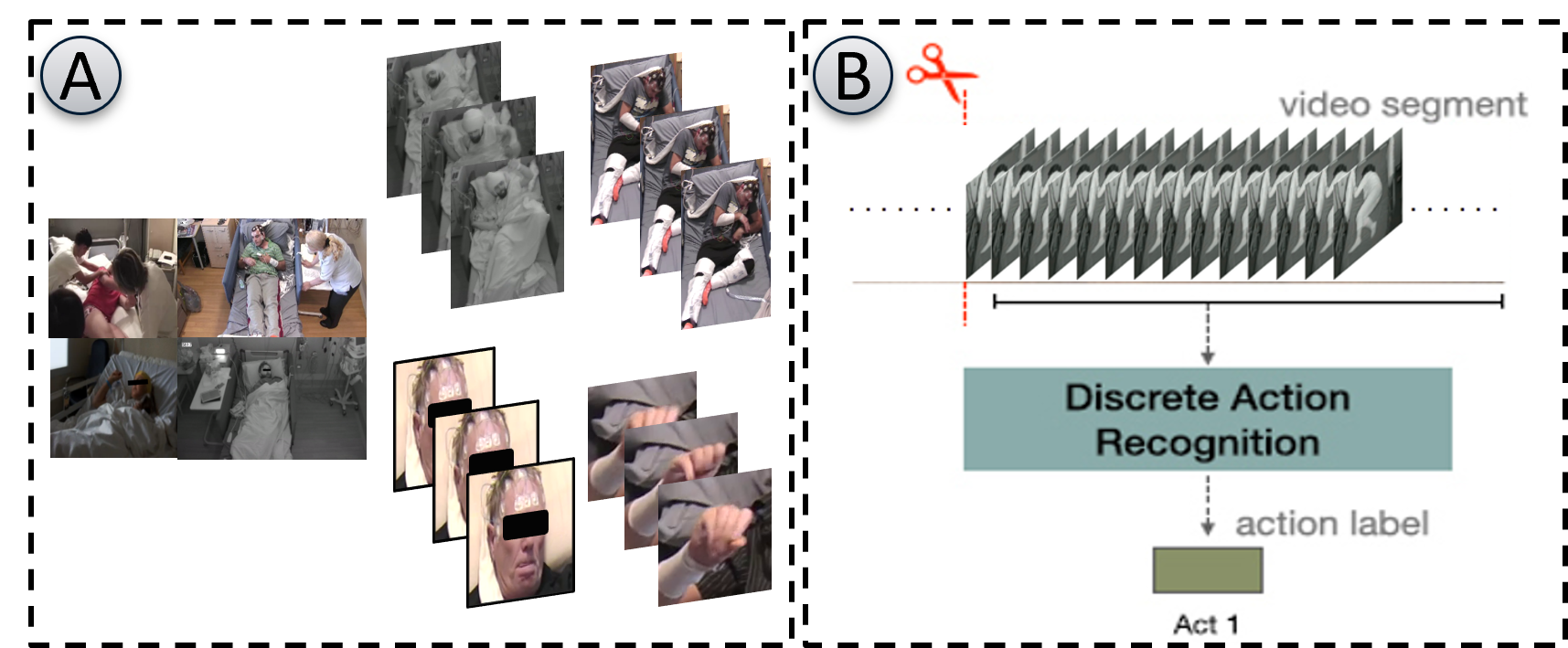}
\vspace{-3pt}
   \caption{Overview of frame-based human action recognition.
   \textbf{A.} Motion capture: Region of interest detection such as the body, face, or hands.
   \textbf{B.} Spatio-temporal representations are learned from sequential frames. CNN-RNN-based, 3D-CNN-based and Transformer-based methods are all commonly used with this type of approach.
   }
\label{fig:HAR_approaches2}
\vspace{-8pt}
\end{figure*}

In frame-based action recognition, the focus is on analyzing regions or patches within sequences of images. This approach does not rely on explicit skeletal data but rather on the visual information captured in the frames, including the spatial layout and temporal changes in these regions~\cite{sun2022huamn}. This framework is illustrated in Figure~\ref{fig:HAR_approaches2}.
As these approaches directly operate on RGB videos, there is a possibility of privacy leakage of sensitive patient data from videos. Moreover, obtaining consent from patients to share their raw RGB video data for inter-cohort validation studies and developing approaches that generalize to large populations becomes challenging. 

\vspace{-6pt}
\subsubsection{Markerless motion capture: region of interest detection}
\label{background:spatiotempo_region}

Object detection is a challenging problem that requires the solution to two main tasks: recognition and localization. Recent years have witnessed remarkable performance gains in object detection, attributable to advancements in deep convolutional neural networks. These detectors have been useful for the identification of key areas in each video such as the patient's body, head, face, mouth, and hands.
Object detectors can be categorized as \textit{anchor-based}, \textit{anchor-free} and \textit{Transformer-based} ~\cite{jiao2019survey,chen20232d}. The core idea of anchor-based models is to introduce a constant set of bounding boxes, referred to as \textit{anchors}, which can be viewed as a set of pre-defined proposals for bounding box regression (methods such as the RCNN series, YOLO series, and SDD series of detectors, \textit{e.g.}~Mask R-CNN~\cite{he2017mask}, Scaled-YOLOv4~\cite{wang2021scaled}, PP-YOLOv2~\cite{huang2021pp}).
Anchor-free methods offer significant promise to cope with extreme variations in object scales and aspect ratios~\cite{ke2020multiple}. Such approaches, for example, can perform object bounding box regression based on anchor points instead of boxes (\textit{i.e.}~the object detection is reformulated as a keypoint localization problem) (key-point-based and anchor-point-based methods, \textit{e.g.}~YOLOX~\cite{ge2021yolox}, PP-YOLOE~\cite{xu2022pp}).
Object detection algorithms based on the Transformer architecture can capture long-range dependencies over the object and easily extract useful global information, which has gradually become a hot research direction in this field (methods include the DETR series and ViT series, \textit{e.g.}~DETR~\cite{carion2020end}, AnchorDETR~\cite{wang2022anchor}, ViT-FRCNN~\cite{beal2020toward}, Swin Transformer~\cite{liu2021swin}, VitDet~\cite{li2022exploring}).
Additional details on object detection algorithms can be found in~\cite{zaidi2022survey,chen20232d}.

\vspace{-6pt}
\subsubsection{Action recognition models (frame-based HAR)}
\label{background:spatiotempo_action}

Frame-based human action recognition involves capturing both spatial and temporal features from video sequences to understand and classify human actions. This task can be executed in either a two-stage or one-stage configuration. In a two-stage setup, spatial features are extracted from each frame, followed by the subsequent modeling of temporal dependencies across the frames. Conversely, in a one-stage configuration, the unified architecture processes all input video frames jointly, capturing both spatial and temporal features. The subsequent descriptions provide more insight into these action recognition models.

\vspace{4pt}
\noindent \textit{CNN-RNN based methods:} 
Several video-based discrete action recognition methods, such as~\cite{simonyan2014two}, have employed a CNN-RNN framework to leverage both spatial and temporal information. Recurrent Neural Networks (RNNs) are widely used to analyze temporal data through the recurrent connections in their hidden layers, and in CNN-RNN frameworks they simply operate over a sequence of embeddings extracted frame-wise with a CNN. An example of this approach is introduced in ~\cite{wan2020action}, where a two-stream convolutional network with long-short-term spatio-temporal features is utilized for human action recognition.

\vspace{4pt}
\noindent \textit{3D-CNN-based methods:} 
Many researchers have extended 2D CNNs to 3D structures to simultaneously model the spatial and temporal information in videos which is crucial for HAR. As one of the earliest works, Ji et al.~\cite{ji20123d} introduced a novel approach to address the limitations of traditional 2D CNNs for recognizing human actions in videos. By incorporating the temporal dimension, the proposed 3D CNN model captures motion information and improves the accuracy of action recognition.
Wang et al.~\cite{wang2017two} integrated a two-stream 3D CNN with an LSTM model to capture long-range temporal dependencies. 
While there are many examples of 3D-CNNs, the temporal structures of 3D convolutional networks are highly inflexible. These networks only accept a set number of frames as input, for example, Ji et al.~\cite{ji20123d} limits it to just 7 frames. Furthermore, setting an appropriate temporal span is challenging given that macro movements for different actions and different body parts have different speeds and durations.

\vspace{4pt}
\noindent \textit{Transformer-based methods:} 
As previously noted, 3D CNN approaches have several drawbacks that limit their applicability when dealing with complex actions of variable durations. Transformer-based approaches bridge this gap by providing a unique solution. The Transformer model can process sequential data in parallel by introducing positional embeddings. 
An early transformer-based method~\cite{girdhar2019video}, Video Action Transformer Network (VATN), consists of two main components: the Spatial Transformer and the Temporal Transformer. 
The learned transformers allow the network to selectively attend to important features, resulting in improved action recognition accuracy.
Recently, the MTV~\cite{yan2022multiview} framework combines multiple views of a video sequence to capture comprehensive spatial and temporal information. This introduces a fusion mechanism that combines the outputs of the transformers. Additionally, ~\cite{feichtenhofer2022masked} introduces a framework that utilizes masked auto-encoders to capture spatial and temporal dependencies in video data. This approach offers a valuable mechanism for leveraging masked auto-encoders with video data to enhance the understanding and recognition of intricate video patterns.
We refer keen readers to~\cite{HerathHP16} and~\cite{ulhaq2022vision} for a more detailed analysis of action recognition methods.


\section{Case studies of vision-based approaches in Epilepsy}
\label{litreview}

For over a decade, research efforts have been dedicated to exploiting the benefits of computer vision for seizure semiology analysis~\cite{pediaditis2010vision,ahmedt2017automated}. Recent approaches leverage the synergy of computer vision and deep learning techniques for motion detection and action recognition. Compared to traditional machine learning approaches, deep learning techniques stand out for their capacity to autonomously learn spatio-temporal feature representations from data, obviating the need for hand-designed features~\cite{ahmedt2017automated}. While the natural inclination to employ deep learning in seizure video analysis is evident, studies in this domain remain relatively scarce compared to those focused on neuroimaging and neuronal activity~\cite{roy2019deep,lhatoo2020big}. An overview of existing vision-based applications for seizure video analysis is detailed in Table~\ref{table:semiology}.
In addition to video-based approaches, there have been contributions focused on numerical representations of seizure descriptors, and textual descriptions detailing the semiology of seizures. These numerical representations, when combined with AI, show promise in localizing the origin of seizures~\cite{mora2022nlp}. However, a limitation persists, as the text and description remain subjective to expert interpretation.

\begin{table*}[t]
\caption{Chronological benchmark of vision-based approaches for semiology analysis (marker-free systems). For pre-2016 works, please refer to our previous review paper~\cite{ahmedt2017automated}.}
\centering
\label{table:semiology}
\resizebox{0.99\textwidth}{!}{%
\begin{tabular}{l l l l c l}
\toprule
\textbf{Signal} & \textbf{Application (Classes)} & \textbf{Target} & \textbf{Authors} & \textbf{Year} & \textbf{Dataset (Location recorded)}\\
%
\midrule
RGB & Detection (FLE, TLE) & Face & Hou et al.~\cite{hou2023artificial}
    & 2023 &  Private (Taipei Veterans General Hospital, Taipei) \\ 
RGB $*$ & Early Detection (TCSs)  & Body & Mehta et al.~\cite{mehta2023privacy}
    & 2023 &  Private (Hospital-based Video EEG Monitoring, not provided) + \href{https://github.com/fepegar/gestures-miccai-2021}{GESTURES}~\cite{perez2021transfer} \\ 
RGB & Classification (SHE, DOA)  & Body & Moro et al.~\cite{moro2023automatic}
    & 2023 &  Private (Sleep Medicine Center, Niguarda Hospital, Italy) \\ 
RGB $*$ & Classification (GTCSs,TCS)  & Body &  Gar{\c{c}}{\~a}o et al.~\cite{garccao2023novel}
    & 2023 &  Private (Hospital de Santa Maria (HSM) and Hospital de Egas Moniz, Portugal) \\
RGB & Detection (Epileptic spasms)  & Body &  Eguchi et al.~\cite{eguchi2023video}
    & 2023 &  Private (Hokkaido University Hospital, Japan) \\
RGB & Classification (Seizures$\star$) $\S$   & Body &  Ojanen et al.~\cite{ojanen2023automatic}
    & 2023 &  Private (Tampere University Hospital, Finland) \\
RGB+IR & Detection (Nocturnal seizures) $\S$ & Body & Lennard et al.~\cite{lennard2023improving}
    & 2023 &  Private (Cornwall, Uk) \\
RGB+IR & Detection (ES$\star$, PNES) $\S$ & Body & Rai et al.~\cite{rai2023automated}
    & 2023 &  Private (Danish Epilepsy Center and Aarhus University Hospital, Denmark) \\
RGB & Detection (Nocturnal seizures$\star$) $\S$   & Body &  Peltola et al.~\cite{peltola2022semiautomated}
    & 2022 &  Private (Danish Epilepsy Center and Aarhus University Hospital, Denmark) \\
RGB+IR & Detection (Nocturnal TCSs) $\S$  & Body &  Armand Larsen et al.~\cite{armand2022automated}
    & 2022 &  Private (Danish Epilepsy Center and Aarhus University Hospital, Denmark) \\
RGB & Classification (ES, PNES)  & Body/Face & Hou et al.~\cite{hou2022self}
    & 2022 &  Private (Timone University Hospital, France) \\ 
RGB & Classification (Dystonia, Emotion) & Body/Face & Hou et al.~\cite{hou2022automated}
    & 2022 &  Private (Timone University Hospital, France) \\ 
IR+RGB-D & Classification (FLE, TLE, NS)  & Body & Kar{\'a}csony et al.~\cite{karacsony2022novel}
    & 2022 &  Private - 3Dvideo-EEG (University of Munich, Germany) \\ 
RGB & Classification (Gelastic, Dacrystic, NS)  & Face & Pothula et al.~\cite{pothula2022real} 
    & 2022 &  Customised (Public Youtube videos) \\ 
RGB & Detection (Rolandic Epilepsy) & Body & Wu et al.~\cite{wu2022multi} 
    & 2022 &  Private (Children’s Hospital, Zhejiang University School of Medicine, China) \\
RGB & Classification (ES, PNES) & Body/Face & Hou et al.~\cite{hou2021multi} 
    & 2021 &  Private (Timone University Hospital, France) \\
%
%
RGB & Classification (FOS, TCS) & Body & P{\'e}rez-Garc{\'\i}a et al.~\cite{perez2021transfer} 
    & 2021 & \href{https://github.com/fepegar/gestures-miccai-2021}{GESTURES} (National Hospital for Neurology and Neurosurgery, UK) \\
RGB & Detection (GTCS) & Body & Yang et al.~\cite{yang2021video} 
    & 2021 &  Private (Boston Children's Hospital, US) \\
RGB & Detection (Myoclonic jerks$\ddag$) & Body & Hypp{\"o}nen et al.~\cite{hypponen2020automatic} 
    & 2020 &  Private (Kuopio Epilepsy Center Helsinki, Finland) \\
IR & Classification (TLE, FLE) & Body & Kar{\'a}csony et al.~\cite{karacsony2020deep} 
    & 2020 &  Private - 3Dvideo-EEG~\cite{cunha2016neurokinect,patricia2018neurokinect} (University of Munich, Germany) \\
RGB $*$ & Detection (Nocturnal seizures$\star$) & Body & van Westrhenen et al.~\cite{van2020automated,kalitzin2012automatic} 
    & 2020 &  Private (Children monitored in their home or in residential care) \\
RGB & Semiology Evolution (PE) & Body/Face & Hou et al.~\cite{hou2020rhythmic}
    & 2020 &  Private (Timone University Hospital, France) \\
IR+RGB-D & Classification (TLE, ETLE) & Body & Maia et al.~\cite{maia2019epileptic} 
    & 2019 &  Private - 3DVideo-EEG~\cite{cunha2016neurokinect,patricia2018neurokinect} (University of Munich, Germany) \\    
RGB & Semiology Evolution (MTLE, ETLE) & Face/Hands & Ahmedt-Aristizabal et al.~\cite{ahmedt2019motion} 
    & 2019 &  Private (Mater Hospital Neuroscience Centre, Australia) \\  
RGB & Classification (MTLE, ETLE) & Face (Mouth) & Ahmedt-Aristizabal et al.~\cite{ahmedt2019vision} 
    & 2019 &  Private (Mater Hospital Neuroscience Centre, Australia) \\  
RGB & Detection Aberrant behavior (MTLE, ETLE) & Body & Ahmedt-Aristizabal et al.~\cite{ahmedt2019aberrant} 
    & 2019 &  Private (Mater Hospital Neuroscience Centre, Australia) \\  
RGB & Classification (MTLE, ETLE, FND) & Body & Ahmedt-Aristizabal et al.~\cite{ahmedt2019understanding} 
    & 2019 &  Private (Mater Hospital Neuroscience Centre, Australia) \\    
RGB & Classification (MTLE/ETLE) & Hands & Pemasiri et al.~\cite{pemasiri2019semantic} 
    & 2019 &  Private (Mater Hospital Neuroscience Centre, Australia) \\
RGB & Detection (Unclear) & Body & Fang et al.~\cite{fang2018spatial} 
    & 2018 &  Information not provided \\
RGB & Classification (MTLE/ETLE) & Body/Face/Hands & Ahmedt-Aristizabal et al.~\cite{ahmedt2018hierarchical} 
    & 2018 &  Private (Mater Hospital Neuroscience Centre, Australia) \\
RGB & Classification (MTLE/ETLE) & Body/Face & Ahmedt-Aristizabal et al.~\cite{ahmedt2018deep} 
    & 2018 &  Private (Mater Hospital Neuroscience Centre, Australia) \\
RGB & Detection (MTLE) & Face & Ahmedt-Aristizabal et al.~\cite{ahmedt2018deep_face} 
    & 2018 &  Private (Mater Hospital Neuroscience Centre, Australia) \\
RGB $*$ & Detection (Nocturnal seizures$\star$) & Body & Geertsema et al.~\cite{geertsema2018automated,kalitzin2012automatic} 
    & 2018 &  Private (University Medical Center Utrecht, Netherlands) \\
IR+RGB-D & Detection ($\dag$) & Body & Achilles et al.~\cite{achilles2018convolutional} 
    & 2018 &  Private (University of Munich, Germany) \\
RGB & Detection (Nocturnal convulsive) & Body & Aghaei et al.~\cite{aghaei2017epileptic} 
    & 2017 &  Private (ChDetecildren's Hospital Medical Center of Tehran, Iran) \\
RGB-D & Classification (TLE/ETLE) & Upper Limbs/Head & Cunha et al.~\cite{cunha2016neurokinect} 
    & 2016 &  Private - 3DVideo-EEG (University of Munich, Germany) \\
\bottomrule
\multicolumn{6}{p{800pt}}
{
Red Green Blue (RGB);
Red Green Blue-Depth (RGB-D);
Infra red (IR);
Temporal Lobe Epilepsy (TLE);
Extratemporal Lobe Epilepsy (ETLE);
Mesial Temporal Lobe Epilepsy (MTLE);
Functional neurological disorders (FND);
Prefrontal Epilepsy (PE);
Focal Onset Seizure (FOS);
Tonic-clonic seizure (TCS);
Psychogenic Non-Epileptic Seizure (PNES);
Sleep-related hypermotor Epilepsy (SHE);
Disorders of arousal (DOA);
Generalized tonic-clonic seizure (GTCS);
Epileptic Seizures (ES);
Non Seizure (NS);
$*$ vector-based representation of body movements (\textit{e.g.} OpticalFlow)
$\star$ Tonic-clonic, tonic, hyperkinetic seizures;
$\dag$ clonic, tonic, automotor seizures;
$\ddag$ prediction unified myoclonus rating scale (UMRS);
$\S$ CE-marked Nelli\textregistered hybrid system produced by \href{https://neuroeventlabs.com/for-patients}{Neuro Events laboratory}.
}
\end{tabular}}
\vspace{-9pt}
\end{table*}

\vspace{-6pt}
\subsection{Applications and model frameworks}
\label{litreview_related}

Initial studies have primarily employed hand-designed features based on motion-strength, motion-trajectory, or average differential luminance signals in video segments to develop automated seizure detection systems. For example, hand-crafted features are extracted based on patient motion trajectories by attaching infrared reflective markers to specific body key points and by exploiting RGB-D sensors (\textit{e.g.}~the Microsoft Kinect camera) to distinguish Temporal Lobe Epilepsy (TLE) and Extratemporal Lobe Epilepsy (ETLE)~\cite{cunha2016neurokinect}. 
Aghaei et al.~\cite{aghaei2017epileptic} proposed a multimodal seizure detection approach to detect nocturnal convulsive seizures by combining hand-crafted features from video and EEG signals, achieving a higher accuracy than using either modality alone. 
Similarly, spatio-temporal interest points are extracted from video sequences to construct a histogram of word frequency features for video representation to detect Rolandic Epilepsy in~\cite{wu2022multi}. 
Other works have focused on nocturnal convulsive seizures using hand-crafted detection algorithms~\cite{geertsema2018automated,van2020automated} by exploiting optical flow calculations and the reconstruction vector field of velocities~\cite{kalitzin2012automatic}. The algorithm quantifies the oscillatory movements seen as vibrations during the tonic phase, and clonic movements in the clonic phase. 
Recently, Gar{\c{c}}{\~a}o et al.~\cite{garccao2023novel} proposed a privacy-preserving video-based seizure-detection and registration method to classify generalized tonic–clonic and focal to bilateral tonic–clonic seizures. This approach is based on optical flow, dimensionality reduction with principal component analysis, seizure movement isolation with independent component analysis, and machine learning classification. 
However, these approaches are limited in performance due to their inability to generalize to changing clinical environments (\textit{e.g.}~changes to subject orientation and distance, dealing with multiple individuals in the field of view, accounting for lighting and furniture changes, motion blur, variable frame rate) and their reliance on handcrafted features.

\vspace{4pt}
\noindent \textit{Seizure semiology detection:} 
Within the landscape of seizure detection, several deep learning models have emerged to improve the accuracy and objectivity on the task at hand. 
Achilles et al.~\cite{achilles2018convolutional} introduced a system that utilizes CNNs on streams from a combined depth and infrared (IR) sensor for seizure detection, outperforming traditional methods like using Histogram of Oriented Gradients features with a Support Vector Machine classifier.
Ahmedt-Aristizabal et al.~\cite{ahmedt2018deep_face} showcased the effectiveness of quantitative facial expression analysis based on deep learning in distinguishing facial semiology from patients with Mesial Temporal Lobe Epilepsy (MTLE) during routine monitoring. 
Hypp{\"o}nen et al.~\cite{hypponen2020automatic} developed an objective myoclonus quantification methods for myoclonus Epilepsy type 1 (EPM1), obtaining a myoclonic jerk score from video recordings using estimated human body key points.
In pediatric patient data, Yang et al.~\cite{yang2021video} employed a CNN-LSTM pipeline to detect generalized tonic-clonic seizures (GTCSs), demonstrating superior performance compared to frame classification analysis (\textit{i.e.}~without considering the temporal dynamics).
Eguchi et al.~\cite{eguchi2023video}, using a similar framework to detect epileptic spasms from long-term video EEG, observed false positives due to voluntary patient movement, such as arm swinging or touching the face, emphasizing the challenges in distinguishing relevant movements.
Hou et al.~\cite{hou2023artificial} explored AI-based face transformation modules, such as a cartoonization, for deidentification and semiology preservation when modelling epileptic seizures across various brain regions, including frontal lobe, temporal lobe, occipital lobe, frontotemporal and temporoparietal regions. While promising, the study revealed a trade-off between deidentification and the loss of clinical details, highlighting the need for specialized face models to capture the heterogeneous nature of facial semiology.

For early detection, Fang et al.~\cite{fang2018spatial} proposed a spatio-temporal gated recurrent unit convolutional neural network capable of not only early detection, but also of localizing the onset time of epileptic seizures. However, the specific seizure types considered and whether the system detects the full expression of the semiology or just the seizure onset remain unclear. 
In a recent study, Mehta et al.~\cite{mehta2023privacy} demonstrated the feasibility of detecting tonic-clonic seizures (TCS) during their progression based on optical flow data. The approach utilized knowledge distillation and transformer-based approaches, showcasing a novel direction for early detection while addressing privacy concerns.

The creation of devices capable of objectively recording seizure counts and characteristics holds immense value, not only for seizures occurring during sleep but also for those manifesting while awake. There has been an increasing interest in the analysis of extensive video data collected at home, particularly for understanding seizure semiology in patients using multiple recording modalities~\cite{amin2021value}. 
A notable example is the Nelli\textregistered hybrid system~\cite{basnyat2022clinical}, a semi-automatic seizure monitoring platform utilizing audio/video inputs. This hybrid system employs computer vision and machine learning to identify kinematic data associated with seizures. Human experts then visually assess these identified events. The utility for reviewing nocturnal video recordings and motor seizures has been demonstrated~\cite{armand2022automated,rai2023automated,lennard2023improving}. 
While Nelli demonstrates proficiency in detecting subtle motor seizures, a study noted a lower accuracy in classifying more discrete motor seizures such as myoclonic jerks, short tonic seizures, and epileptic spasms~\cite{peltola2022semiautomated}. The system has received endorsement for clinical use in Finland from a government-established body, the National Coordinating Group for Drug-resistant Epilepsy.

\vspace{4pt}
\noindent \textit{Seizure semiology classification:}
Understanding the evolution of automated pipelines in the context of video analysis for seizure event classification is crucial. While action recognition requires extracting spatio-temporal features of movements for the classification of epilepsy types, some works overlook the temporal information within seizures, treating the task as an image classification problem. For example, Maia et al.~\cite{maia2019epileptic} used an InceptionV3 feature extractor to extract features for all frames, which were subsequently concatenated and fed to a multilayer perceptron, to distinguish between TLE and ETLE within IR seizure videos. Similarly, Pothula et al.~\cite{pothula2022real} differentiated between gelastic and dacrystic facial seizures using a feature extraction layer that recognizes the patient's emotion through a facial emotion recognition model and classification layer (Random Forest) that received concatenated features. These approaches ignore vital temporal information by simply concatenating features across frames, and thus fail to understand how behavior is changing over time (a key aspect for optimal seizure semiology analysis).
%
In contrast, Kar{\'a}csony et al.~\cite{karacsony2020deep} employed a 3D-CNN (I3D) and an LSTM to extract spatio-temporal features to distinguish TLE and FLE using IR data. Then, the same authors in~\cite{karacsony2022novel}, extended the architecture to 3 classes including a non-seizure class, and introduced a new pre-processing pipeline, utilizing depth cropping to improve classification performance. 
Moro et al.~\cite{moro2023automatic} extended this architecture to differentiate between non-hyperkinetic seizures and sleep-related paroxysmal events, utilizing a majority voting scheme for the entire video prediction.

Approaches focusing on appearance from the full body as a region of interest expose a limitation in evaluating specific semiology. For example, missing important clinical signs such as mouth automatisms (\textit{e.g.}~chewing, ictal pouting, and smacking) and hand and finger semiology (\textit{e.g.}~waving, snapping finger and tapping) due to the limited resolution that renders these semiology invisible.
Ahmedt-Aristizabal et al.~\cite{ahmedt2018hierarchical} introduced a hierarchical approach, analyzing facial expressions, head and upper limb motions, and hand and finger movements to classify MTLE and ETLE. A hierarchical approach tackles complex problems by reducing them to a smaller set of interrelated sub-systems, thus the authors did not perform decision fusion, emphasizing the importance of each clinical sign based on the development and sequence of multiple semiological features.
Ahmedt-Aristizabal et al.~\cite{ahmedt2019vision} introduced 3D face models that retain rich information about the shape and appearance to evaluate mouth and cheek motions. From the resultant 3D mesh of the face, spatial features are fed to an LSTM to extract temporal relations between sequences to classify MTLE and ETLE.
Pemasiri et al.~\cite{pemasiri2019semantic} proposed a unified semantic segmentation approach of body parts that precisely refines hand boundaries. This step is critical to filtering out motion related to bedding and monitoring equipment, which is irrelevant to hand semiology. The extraction of hand features follows the methodology in~\cite{ahmedt2018hierarchical}. First, spatial features are extracted from a semantic segmentation architecture, then dynamic variations of each hand are captured with an LSTM to distinguish between MTLE and ETLE. 
Most works classify epilepsy types from short video clips (snippet-level classification), obtaining a subject-level prediction by averaging all snippet-level predictions. To avoid potential misclassifications, Perez et al.~\cite{perez2021transfer} used a spatio-temporal network to distinguish focal onset seizures (FOSs) from focal to bilateral TCSs. The authors used temporal segment networks to capture semiological features across the entirety of the seizure, and an RNN as a consensus function to learn seizure-level representations from the sequence of features.
Recently Ojanen et al.~\cite{ojanen2023automatic} demonstrated the potential of the Nelli system discussed previously to classify tonic, tonic-clonic, and hyperkinetic seizures by utilizing motion and oscillation signal profiles.

Studies have explored multi-stream frameworks to analyze clinical manifestations from the face, head and upper limbs movements.
Ahmedt-Aristizabal et al.~\cite{ahmedt2018deep} proposed a multi-framework fusion approach for facial and pose dynamics, achieving positive results in identifying semiological features associated with focal epilepsies, such MTLE and ETLE. The fusion process aims to extract useful information from a set of different input modalities and merge them in such a way that allows increasingly accurate and robust decisions to be made.
Similarly, Hou et al.~\cite{hou2022automated} adopted a fusion framework for discriminating between features of dystonia and emotion in hyperkinetic seizure videos using multi-stream information from keypoints and appearance. Rather than using a standard combination framework such as CNN-RNN architectures, the authors adopted GCNs for seizure analysis. While fusion approaches can improve classification accuracy, challenges arise when patients do not experience face and body semiology in all seizures, or when videos only allow extracting facial features due to occlusions.

\vspace{4pt}
\noindent \textit{Distinguishing epileptic seizures from non-epileptic seizures:} 
A main differential diagnosis of epileptic seizures (ES) is psychogenic non-epileptic seizures (PNES, also known as functional or dissociative seizures, part of the spectrum of functional neurological disorder, FND). PNES have a non-epileptic cause thought to involve psychological and as yet poorly understood neurological mechanisms~\cite{benbadis2000estimate}. Given that both ES and PNES can exhibit rhythmic movements, misdiagnosis is common due to overlapping clinical features observed during these 2 distinct seizure types. It is crucial to differentiate between the two, as their clinical management differs, and misdiagnosis may result in unnecessary treatment and associated complications. We refer in this section to both diagnostic groups, epileptic seizures (ES) and FND, under the common label of seizure disorders.
In the study by Ahmedt-Aristizabal et. al.~\cite{ahmedt2019understanding}, skeleton-based and frame-based frameworks were compared, with both providing effectiveness in distinguishing between seizure disorders. The authors quantified semiology using either a fusion of reference points and flow fields or by analyzing the entire body simultaneously. While this design is valuable for evaluating PNES, the region-based approach raises a clinical limitation regarding the observation of isolated semiology.
In another work, Hou et al.~\cite{hou2021multi} demonstrated a method to process the appearance and keypoint information for both pose and face streams. They introduced an adaptive GCN where the graph's topology could be learned on detected body joints and facial landmarks for seizure classification. Knowledge distillation was also adopted to regulate the keypoint features learned by the adaptive GCN. However, including joint information may render the system vulnerable in conditions where joint estimation is poorly performed.
In a subsequent work~\cite{hou2022self}, the same authors exploited appearance with a Transformer-based framework. This model was pre-trained on large, unlabeled clinical videos (contextual videos) and then fine-tuned to distinguish between ES and PNES.

\vspace{4pt}
\noindent \textit{Seizure anomaly detection:} 
In existing detection and multi-framework strategies for classification, the sensitivity to potential aberrant or unusual semiology poses a considerable challenge. The core task in this context is the identification of data samples that deviate from the overall data distribution, a principal task in anomaly detection. Anomaly detection methods play a crucial role in pinpointing interesting, concerning, or unknown events by leveraging past patient cases stored in health records. Despite the straightforward definition, the implementation of anomaly detection faces challenges, primarily from the inconsistent behavior of different anomalies and the lack of a constant definition for what constitutes an anomaly~\cite{fernando2021deep}. As a result, effective learning with high modeling capacity is required to segregate anomalous samples from normal data, a task that has been insufficiently addressed in the existing vision-based seizure semiology research.
Ahmedt-Aristizabal et al.~\cite{ahmedt2019aberrant} introduced a system designed to identify aberrant epileptic seizures, providing clinicians with alerts regarding the occurrence of unusual events that deviate from a pre-learned database of known semiology. This approach clusters epileptic seizures into a best-fit model, using a template of stereotypical behaviors of MTLE and ETLE in the form of libraries. These libraries store feature representations of the motion exhibited by patients during a recorded seizure. This approach enables users to determine if the semiology from a new patient aligns with the learned information, aiding in the identification of dissimilar findings or aberrant semiology.

\vspace{4pt}
\noindent \textit{Seizure evolution analysis:} 
Recent advances in video analytics have been useful in capturing and quantifying epileptic seizures. However, a critical aspect that has often been overlooked is the representation of the evolving semiology, specifically tracking the stepwise progression of clinical features. Many studies focused on the detection and categorization of movement patterns, neglecting the quantification of multi-segmental rhythmic behaviors and the identification of certain movements, including the order in which they occurred. Understanding the electroclinical patterns of a seizure requires a spatio-temporal profile to be established, allowing for the evaluation of the seizure's origin and propagation patterns~\cite{chauvel2014emergence}.
Approaches incorporating statistical information offer quantitative movement parameters by considering the entire duration of semiology~\cite{cunha2016neurokinect}. Previous deep learning frameworks for HAR provide a single result reflecting the classifier’s decision and cannot convey the dynamic changes observed.
Ahmedt-Aristizabal et al.~\cite{ahmedt2019motion} developed a system that captures the motion dynamics of body, face and hand semiology, enabling the visualization and analysis of dynamic changes over an entire seizure. This system utilizes a compact image representation known as a motion signature, which captures the spatial location of semiology and the temporal relation between frames. It can also correlate different types of semiology when a patient experiences multiple semiologies simultaneously, thus displaying the order of signals as a stepwise progression.
Hou et al.~\cite{hou2020rhythmic} reported that automated video analysis enabled the characterization of rhythmic body movements as a stereotyped expression of frontal seizure semiology. The authors documented time-evolving frequencies as a foundation for studying the correlation between the spectrum of EEG and the frequency of head movements. 
Developing a system that provides a continuous flow of signals, highlighting changes in semiology and common clinical events, could be a powerful tool to support seizure diagnosis and classification.
Indeed, such developments applied to larger clinical datasets could shed new light on currently unknown neural mechanisms underlying the behavioral expression of seizures.

\begin{table*}[!t]
\caption{Data driven approaches adopted in Epilepsy for semiology analysis.}
\centering
\label{table:CV_techniques}
\resizebox{0.98\textwidth}{!}{%
\begin{tabular}{
>{\raggedright\arraybackslash}p{3.2cm}|
>{\raggedright\arraybackslash}p{11cm}|
>{\raggedright\arraybackslash}p{3cm}}
\toprule
\textbf{Approach}                          & \textbf{Deep learning Methods}          &  \textbf{Authors} \\
\midrule
\multicolumn{3}{l}{\textbf{MoCap: keypoints/mesh and region of interest}} \\  \cline{1-3}
\multirow{3}{3.2cm}{Patient Detection}     & Faster R-CNN~\cite{ren2015faster}       & ~\cite{ahmedt2018hierarchical}     \\ 
                                           & Mask R-CNN~\cite{he2017mask}            & ~\cite{ahmedt2019understanding,ahmedt2019motion,maia2019epileptic,karacsony2020deep,karacsony2022novel}  \\ 
                                           & SSD~\cite{liu2016ssd}                   & ~\cite{hou2021multi,hou2022automated,hou2022self}     \\ \cline{1-3}
\multirow{5}{3.2cm}{Face Detection \& Tracking} & Faster R-CNN~\cite{ren2015faster,jiang2017face}      & ~\cite{ahmedt2018deep_face,ahmedt2018deep} \\ 
                                           & Faster R-CNN~\cite{ren2015faster,jiang2017face} + Erdos-Renyi clustering~\cite{jin2017end}  & ~\cite{ahmedt2018hierarchical} \\
                                           & Finding Tiny faces~\cite{hu2017finding} + Erdos-Renyi clustering~\cite{jin2017end}          & ~\cite{ahmedt2019vision}    \\ 
                                           & Finding Tiny faces~\cite{hu2017finding} + DeepSORT~\cite{wojke2017simple}                   & ~\cite{ahmedt2019motion} \\                               
                                           & SSD~\cite{liu2016ssd}                   & ~\cite{hou2020rhythmic,hou2021multi,hou2022automated}     \\ \cline{1-3}
\multirow{3}{3.2cm}{Face landmark}           & TCDCN (2D)~\cite{zhang2015learning}           & ~\cite{ahmedt2018deep_face} \\  
                                           & FaceAlignment (2D)~\cite{bulat2017far}        & ~\cite{hou2021multi} \\
                                           & PRNet (3D)~\cite{feng2018joint}               & ~\cite{ahmedt2019vision}     \\ \cline{1-3}
\multirow{2}{3.2cm}{Face transformation}   & FaceSwap~\cite{xu2022mobilefaceswap}          & ~\cite{hou2023artificial} \\  
                                           & VToonify~\cite{yang2022vtoonify}              & ~\cite{hou2023artificial}     \\ \cline{1-3}
\multirow{3}{3.2cm}{Hands Detection \& Tracking} & Multiview bootstrapping (pose-based)~\cite{simon2017hand}      & ~\cite{ahmedt2018hierarchical}    \\ 
                                           & Detect-and-track (pose-based) ~\cite{girdhar2018detect}        & ~\cite{ahmedt2019motion}     \\ 
                                           & Mask R-CNN~\cite{he2017mask}                                   & ~\cite{pemasiri2019semantic}     \\ \cline{1-3}
\multirow{2}{3.2cm}{Hand landmark}          & Multiview bootstrapping~\cite{simon2017hand}  & ~\cite{ahmedt2018hierarchical}     \\  
                                           &        &      \\ \cline{1-3}
\multirow{6}{3.2cm}{Human pose estimation \& Tracking}   & PAF + CPM (2D pose) ~\cite{cao2017realtime}        & ~\cite{ahmedt2018deep,hypponen2020automatic}   \\  
                                           & KeypointRCNN (2D HPE)~\cite{he2017mask}                       & ~\cite{hou2021multi}     \\   
                                           & Thin-Slicing Network (2D HPE) ~\cite{song2017thin}            & ~\cite{ahmedt2018hierarchical}     \\
                                           & LSTM Pose Machines (2D HPE)~\cite{luo2018lstm}                & ~\cite{ahmedt2019understanding}     \\ 
                                           & 2D pose + 3D estimator (3D HPE)~\cite{chen20173d,zhou2017towards}  & ~\cite{ahmedt2018deep,ahmedt2018hierarchical,hou2021multi}  \\ 
                                           & Sensor-based (Microsoft Kinect camera) (2D/3D HPE)~\cite{zhang2012microsoft,patricia2018neurokinect,karacsony2021deepepil}  & ~\cite{cunha2016neurokinect}  \\ 
\midrule
\multicolumn{3}{l}{\textbf{Video-based seizure analysis}} \\ \cline{1-3}
\multirow{12}{3.2cm}{Seizure detection (Dynamic)} & STIP + FE + KNN                              & ~\cite{aghaei2017epileptic}     \\ 
                                           & STIP + FE + SVM                                  & ~\cite{wu2022multi}     \\
                                           & OpticalFlow~\cite{kalitzin2012automatic} + FE    & ~\cite{geertsema2018automated,van2020automated}     \\ 
                                           & OpticalFlow~\cite{farneback2003two} + FE         & ~\cite{garccao2023novel}     \\
                                           & OpticalFlow~\cite{perez2013robust} + TSN~\cite{wang2018temporal} + ViT~\cite{dosovitskiy2020image} + PKD~\cite{zheng2023egocentric}   & ~\cite{mehta2023privacy}     \\ 
                                           & CNNs + FC                                        & ~\cite{achilles2018convolutional,yang2021video}  \\
                                           & CNN-GRU~\cite{cho2014learning} + (Two Stream CNNs~\cite{simonyan2014two} + TSN~\cite{wang2016temporal}) + FC  & ~\cite{fang2018spatial}  \\
                                           & CNN-LSTM~\cite{donahue2015long,greff2016lstm} + FC & ~\cite{ahmedt2018deep_face,eguchi2023video} \\
                                           & CNN-LSTM + MoCap library + Cosine similarity     & ~\cite{ahmedt2019aberrant} \\ 
                                           & 2D HPE + FE + SVM                                  & ~\cite{ahmedt2018deep_face} \\                                             
                                           & 2D HPE + Optical Flow + FE + Statistics           & ~\cite{hypponen2020automatic} \\
                                           & The Nelli\textregistered hybrid system ~\cite{basnyat2022clinical}  & ~\cite{armand2022automated,peltola2022semiautomated,rai2023automated,lennard2023improving} \\ \cline{1-3}
\multirow{3}{3.2cm}{Seizure classification (Static)}     & InceptionV3~\cite{szegedy2015going} + MLP          & ~\cite{maia2019epileptic}  \\ 
                                           & FER~\cite{ekman1978facial} + RF                    & ~\cite{pothula2022real}    \\ 
                                           &        &      \\  \cline{1-3}
\multirow{3}{3.2cm}{Seizure classification (Skeleton-based HAR)}  & 3D HPE$\star$ + Optical Flow + FE + Statistics & ~\cite{cunha2016neurokinect}     \\  
                                           & 3D HPE + FE + LSTM + FC                        & ~\cite{ahmedt2018deep,ahmedt2018hierarchical,ahmedt2019vision} \\ 
                                           &        &      \\ \cline{1-3}
\multirow{5}{3.2cm}{Seizure classification (Frame-based HAR)}              & CNN-LSTM~\cite{donahue2015long,greff2016lstm} + FC & ~\cite{ahmedt2018deep,ahmedt2018hierarchical,pemasiri2019semantic,ahmedt2019understanding,yang2021video}     \\
                                           & I3D~\cite{carreira2017quo} + LSTM~\cite{greff2016lstm} + FC                                                         & ~\cite{karacsony2020deep,karacsony2021deepepil,karacsony2022novel}  \\
                                           & I3D~\cite{carreira2017quo} + Majority voting                                                   & ~\cite{moro2023automatic}  \\
                                           & STCNN (TSNs)~\cite{wang2018temporal,ghadiyaram2019large} + Bi-LSTM~\cite{singh2016multi} + FC  & ~\cite{perez2021transfer}  \\
                                           & CNN (ResNet-152)~\cite{he2016deep} + Transformer~\cite{vaswani2017attention,wolf2019huggingface} + FC    & ~\cite{hou2022self}     \\  \cline{1-3}
\multirow{4}{3.2cm}{Seizure classification (Hybrid approach HAR)} & \big(2D pose + FE + LSTM~\cite{greff2016lstm}\big) + \big(OpticalFlow~\cite{pathak2017learning} + LSTM\big) + FC      & ~\cite{ahmedt2019understanding} \\
                                           & \big(3D pose + AGCNs~\cite{kipf2016semi,shi2019two} + KD~\cite{hinton2015distilling,pan2020spatio}\big) + \big(R(2+1)D~\cite{tran2018closer} and VGG-19~\cite{simonyan2014very}+ TCNs~\cite{lea2017temporal}\big)  & ~\cite{hou2021multi,hou2022automated}   \\ 
                                           &        &      \\ 
\bottomrule
\multicolumn{3}{p{500pt}}
{
Tasks-constrained deep convolutional network (TCDCN);
Position map Regression Network (PRNet);
Human Pose Estimation (HPE);
Part Affinity Fields (PAF);
Convolution Pose Machines (CPM);
Long Short Term Memories (LSTM);
Feature Engineering (FE) (handcrafted feature extraction);
Spatio-temporal interest points (STIP);
K-Nearest Neighbors (KNN);
Support Vector Machines (SVM);
Convolutional Neural Networks (CNNs);
Fully Connected layer (FC);
Motion Capture (MoCap);
Facial emotion recognition (FER);
Random Forest (RF);
Temporal Segment Networks (TSNs);
Spatio-temporal CNN (STCNN);
Vision Transformer (ViT);
Knowledge Distillation (KD);
Progressive Knowledge Distillation (PKD);
Adaptive Graph Convolutional Networks (AGCNs);
Gated Recurrent Unit (GRU);
Multilayer perceptron (MLP);
Bidirectional LSTM (Bi-LSTM);
$\star$ Human pose detected with depth camera hardware;
}
\end{tabular}}
\vspace{-4pt}
\end{table*}

\vspace{4pt}
\noindent \textit{Summary:} A comprehensive overview of diverse deep learning techniques, their corresponding sources, and the authors from Table~\ref{table:semiology} who have adapted each technique is presented in Table~\ref{table:CV_techniques}. The related works are systematically categorized according to the frameworks discussed in Section~\ref{background:overview}. The categories include, 
seizure detection~\cite{aghaei2017epileptic,wu2022multi,geertsema2018automated,van2020automated,garccao2023novel,achilles2018convolutional,ahmedt2018deep_face,hypponen2020automatic,eguchi2023video,fang2018spatial,mehta2023privacy,ahmedt2019aberrant},
image-based (static) seizure classification~\cite{maia2019epileptic,pothula2022real},
skeleton-based seizure classification~\cite{cunha2016neurokinect,ahmedt2018deep,ahmedt2018hierarchical,ahmedt2019vision},
frame-based seizure classification~\cite{ahmedt2018deep,ahmedt2018hierarchical,pemasiri2019semantic,ahmedt2019understanding,yang2021video,karacsony2020deep,karacsony2021deepepil,karacsony2022novel,moro2023automatic,perez2021transfer,hou2022self}, and
hybrid networks for seizure classification~\cite{ahmedt2019understanding,hou2021multi,hou2022automated}.
This classification facilitates a structured exploration of the diverse methodologies employed across different aspects of seizure analysis within the deep learning paradigm.

\vspace{-6pt}
\subsection{General observations regarding vision-based seizure analysis methods}

While much progress has been made with respect to methods to detect and classify seizures, there are many common limitations and/or oversights with respect to these methods. Given the overview of existing applications in Section \ref{litreview_related}, we make the following observations:

\begin{enumerate}

\item The automated analysis of video-recorded seizures is greatly needed to support the detection and identification of epileptic seizures, as well as to better understand the temporal evolution of seizures, from their clinical onset through to termination. While this task has benefited from the use of advanced computer vision and machine learning algorithms, methods are limited by their use of short video clips rather than the entire video of a seizure. This current approach limits the ability of systems to precisely detect the clinical onset, or capture in fine-detail the propagation of semiology.

\item Improving the spatio-temporal consistency of extracted features in a clinical environment is crucial for improved analysis. This is particularly evident when considering the consistency of the keypoints and regions of interest that are detected across sequential images, a priority that has been overlooked in certain works.

\item For systems to be deployed in a real-world setting it is crucial that practitioners are confident that models can generalize to unseen patients. 
This is particularly important for analyzing complex motor behaviors in seizures, since the overall behavioral repertoire and inter-subject and even intra-subject variability is considerable.
As such, when developing and evaluating systems researchers must ensure separation of patients between training and testing sets, such that all footage of a given patient lies exclusively in either the training or the testing set to evaluate the generalizability of the proposed system. In some works, this data separation was not considered, which could produce misleadingly high performances. The inclusion of subject-specific features may introduce data leaks, boosting performance without a genuine basis for the target classification.

\item One limitation of multi-framework approaches for quantifying and classifying clinical manifestations is their reliance on supervised learning, which assumes the test data originates from one of the training categories. When an input sample does not align with any of the training categories (model uncertainty), the trained model becomes ineffective. This is evident when considering the limited performance when classifying the semiology of unseen patients, as indicated by a low Leave-One-Subject-Out Cross-Validation accuracy. This limitation to analyzing seizure-specific features not represented in the training set raises concerns about a network's generalization capabilities.

\item Further understanding of the dynamic changes in movement frequency and amplitude of stereotyped movements across a range of pathologies could help shed light on possible shared pathophysiological mechanisms. To this end, documenting the kinematic properties of stereotypical movements using automated video analysis could be a useful tool.

\item In the domain of vision-based action recognition, it is crucial to exercise caution when comparing methodologies, as performance metrics are reliant on supervised learning from specific behaviors. Not all semiologies and subjects exhibit uniform characteristics, making direct comparisons between methods difficult. Without a standardized public dataset featuring raw videos for benchmarking, claims of superiority for one method over another risk being misleading. A meticulous approach to comparison involves considering the dataset's composition, the diversity of represented behaviors, and the context of the analysis. It is essential to recognize that each methodology may excel under specific conditions while exhibiting limitations under others.

\item To facilitate fair comparisons, recent releases of public datasets based on optical flow representation of different seizure semiologies~\cite{perez2021transfer,mehta2023privacy}, provide a valuable resource for researchers to conduct standardized evaluations of frame-based HAR approaches. A recent open seizure database that facilitates research into non-EEG seizure detection~\cite{pordoy2023open} (accelerometry, photoplethysmography, and pulse-oximetry) is expected to foster collaborative efforts and advance the field of seizure detection. However, there is an increasing need for datasets that provide more comprehensive visual context, particularly regarding behaviors such as facial expressions and hand movements. Striking a balance between data sharing and privacy preservation is challenging yet essential. Possibly, using facial transformation methods in clinical video datasets could mitigate privacy issues and allow datasets of sufficient size to be assembled~\cite{hou2023artificial}. 

\item In the context of seizure analysis, addressing the challenges associated with the ``black box'' nature of machine learning models is essential. Explainable AI approaches are expected to provide more value for clinicians, fostering trust and understanding in the application of these models in real-world medical scenarios~\cite{kundu2021ai}. However, explainable AI methods must be suitable for use by non-deep learning experts, something that is arguably not the case for present interpretation techniques.

\item Generalisation to unseen subjects has been a consistent challenge when performing medical machine learning tasks due to the high degree of subject variability. The recent emergence of foundation models, large-scale models trained in a self-supervised manner on diverse datasets, offers a potential direction to improve generalization if video-based foundation models can be leveraged as a starting point for model development.

\item Within the broader deep-learning field there has been a trend towards building end-to-end systems, that is a single system that takes an input and produces the desired output with no intermediate steps or additional outputs. This paradigm is less prevalent within Epilepsy analysis, though recent end-to-end transformer-based methods concerned with whole body analysis have been proposed. The lack of end-to-end systems likely stems from a combination of the lack of suitably large datasets to train such systems, a reliance on existing methods (\textit{i.e.}~pose detection) to underpin the methods, and also the lack of interpretability that such end-to-end systems offer. It is also worth considering if such end-to-end methods are even appropriate, as processing a whole-body image in an end-to-end fashion is likely to result in hand and face regions are that too small from which to extract fine motor movements.

\end{enumerate}



\section{Pipeline for quantified analysis of seizure video recordings}
\label{guideline}

\begin{figure*}[!t]
\centering
\includegraphics[width=0.99\linewidth]{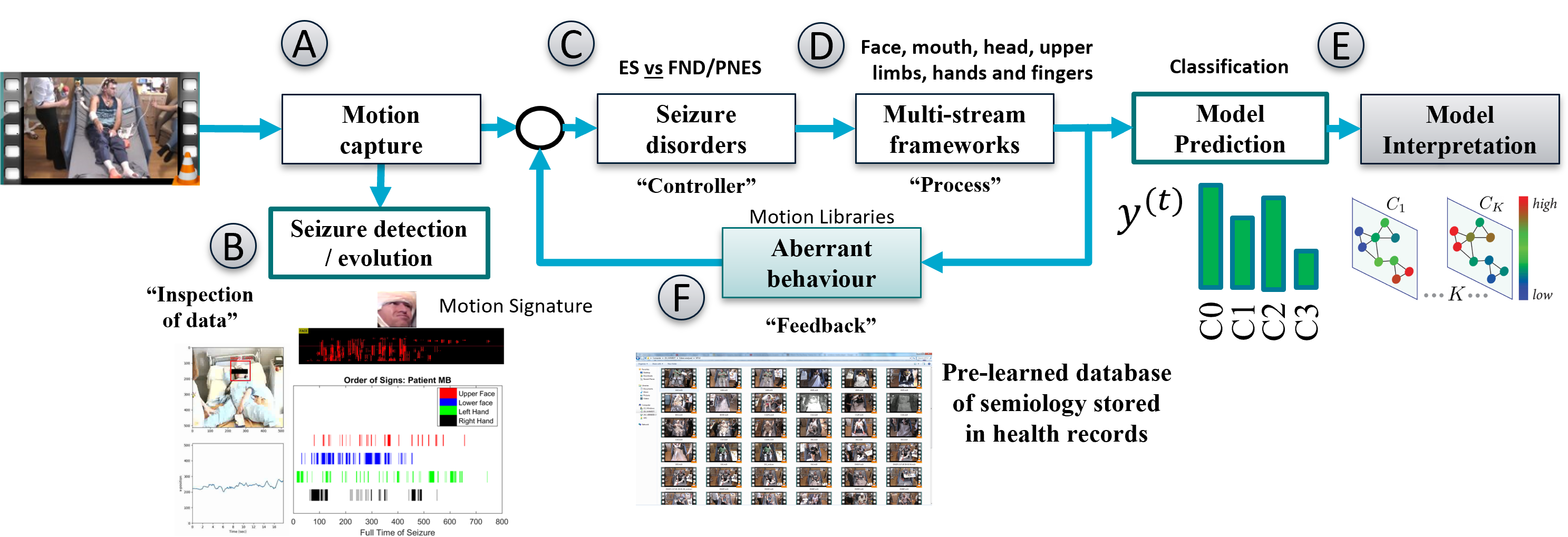}
\vspace{-3pt}
   \caption{Overview of an automatic seizure video evaluation system.
   A potential implementation could be a closed-loop system designed for clinical deployment in the hospital that integrates: 
  \textbf{A.} A flexible and modular markerless motion capture solution,
  \textbf{B.} A method enabling the inspection of the stepwise progression of clinical features and the order of signs,
  \textbf{C.} An analysis of the whole body to distinguish seizure disorders such as PNES,
  \textbf{D.} A flexible and modular multi-stream framework to evaluate isolated semiologies, 
  \textbf{E.} An inspection of the classification results and model uncertainty (model confidence), as well as a set of model-agnostic methods to interpret deep learning methodologies, and
  \textbf{F.} An anomaly detection module to update the system with aberrant video features of semiology.
  ES: Epileptic seizures; FND: functional neurological disorder; PNES: psychogenic non-epileptic seizures.
   }
\label{fig:pipeline}
\vspace{-6pt}
\end{figure*}

In addressing the complexities of vision-based semiology analysis for seizure video recordings, our proposed pipeline introduces a comprehensive methodology comprising distinct modules connected to form an integrated system. This system is unique and provides important supplementary and unbiased data to assess video-recorded seizures through clinical implementation. The proposed ecosystem encompasses key components such as motion capture of keypoints and regions of interest, facilitating the precise extraction of relevant spatial information. Further, motion signatures and time-evolving frequencies of stereotyped movements contribute invaluable insights into patterns that characterize seizure-related behaviors. The pipeline extends to the analysis of seizure disorders and robust seizure detection mechanisms. Multi-stream frameworks have been incorporated for refined seizure classification, ensuring an understanding of varied seizure types. Moreover, an inspection of aberrant behaviors, model prediction verification, and model interpretation modules enhance the interpretability and reliability of the entire system. Such a dynamic and modular pipeline is depicted in Figure~\ref{fig:pipeline}. 
The subsequent subsections provide detailed explanations of each module, elucidating their roles and contributions within this unified framework. 

\vspace{-6pt}
\subsection{Motion capture} 
The motion capture module constitutes the foundational component within the proposed pipeline, and is designed for the detection and representation of motor seizures characterized by their distinctive motion patterns. It employs keypoints and regions of interest to obtain a detailed representation, integrates 3D representations to enhance spatial understanding, and incorporates feature tracking with temporal connections to ensure motion coherence during a seizure. The module also includes scene segmentation techniques to differentiate patient movements from unrelated activities, addressing challenges such as varying camera angles and resolutions. The module extends to include occlusion augmentation during training for improved detection in challenging environments, as well as super-resolution or image enhancement techniques to handle low-resolution images. It incorporates physics-constrained modeling such as biomechanical dynamics methods, and may involve emulation approaches to represent complex dynamics~\cite{harrison2023whole}. 
Despite the motivations outlined in this manuscript and the wealth of related work, the application of deep learning to seizure video analysis remains considerably under-explored. This is evidenced by the adoption of model architectures that are not at the forefront of recent advancements, as indicated in Table~\ref{table:CV_techniques}. In comparison to the rapid progress observed in computer vision research, particularly in recent years, the exploration of deep learning within the domain of seizure video analysis appears to lag significantly representing a noticeable disparity in the advancement of these respective fields.
The flexibility of this module allows for continuous improvement through the integration of advanced computer vision approaches outlined in Section~\ref{background:pose_keypoints} and Section~\ref{background:spatiotempo_region}. This adaptability ensures the module's resilience to technological advancements, maintaining accuracy and robustness in motion detection. Further research opportunities addressing the challenges of motion capture are described in Section~\ref{opp_mocap}, along with privacy preservation solutions for patients in Section~\ref{opp_privacy}.

\vspace{-6pt}
\subsection{Seizure detection and temporal analysis} 
A comprehensive examination of the evolution of video-recorded seizure-related clinical manifestations in patients with Epilepsy is imperative. From an algorithmic perspective, the initial phase involves the detection or identification of a relevant time range for further analysis. The generic model performing this task must be versatile enough to accommodate various seizure types, allowing algorithms to isolate moments of interest and compare extracted features to predefined thresholds. Diverse research and commercial options were documented in Section~\ref{litreview_related}.
The analysis of seizure evolution, aimed at discerning the presence or absence of specific movement features and dynamic changes, is a major component of Epilepsy patient assessment. Seizure manifestations that can be readily detected in present video analyses encompass a spectrum, ranging from repetitive rhythmic movement of trunks, limbs or hands, such as whole body rocking or manipulation of an object; to a more intense presentation characterized by excessive amounts of amplitude, speed, and acceleration~\cite{bonini2014frontal,beniczky2022seizure}. As well as body/limb movement patterns, seizure-related facial changes may bring valuable semiologic information, \textit{e.g.}~muscular contraction, eye deviation or emotional facial expression~\cite{hou2022automated}.
This module, illustrated in Figure~\ref{fig:pipeline}, focuses on the development of a computer-based tool designed to analyze the stepwise progression (or simultaneous occurrence) of clinical features. This tool has the capability to discern dominant features, categorizing them based on speed, periodicity, the frequency of events, and differences in motions. Moreover, the module explores the correlation of motion across different body parts, given that epileptic seizures often transition from one clinical manifestation to another over a certain timescale that might help predict seizure propagation. 
The incorporation of a motion signature or semiological fingerprint proves to be valuable when scrutinizing videos, presenting semiology as a flow of signs~\cite{ahmedt2019motion}. To enhance the interpretability of the identified kinematic data, human experts can visually assess the time-evolving frequency properties of seizures~\cite{hou2020rhythmic} which may support an understanding of the mechanisms underlying such behaviors. Additional details regarding an automated fine-grained action analysis are discussed in Section~\ref{opp_fine-grained}.
%

\vspace{-6pt}
\subsection{Analysis and detection of seizure disorders} 
Prior to the execution of vision-based semiology classification for epilepsies, the ``Controller'' in the system will exclusively provide to the ``Process'' module seizure disorders categorized as epileptic seizures. As discussed previously in Section~\ref{litreview_related}, seizure semiology
is an essential component for patient monitoring and can differentiate between epileptic and non-epileptic seizures. This module focuses on identifying PNES or FND, which lack the characteristic changes in cortical activity associated with epilepsy, sometimes contributing to misdiagnosis.
To address this, the module incorporates end-to-end frameworks that analyze the entire body simultaneously, adopting proper seizure-level prediction~\cite{perez2021transfer} capturing semiological features across the entirety of arbitrarily long seizure, rather than averaging all snippet-level predictions from short clips. Averaging predictions is effective when the same action occurs along most of the seizure video duration, but given the variable propagation of semiology, this is not the typical case. This module utilizes temporal segment networks to split videos of any duration into $n$ non-overlapping segments and extract spatial features from input frames~\cite{perez2021transfer}, as opposed to traditional CNNs~\cite{ahmedt2019understanding}. Subsequently, the module leverages transformers to effectively learn temporal relations between the extracted spatial features of the seizure patterns, as demonstrated in previous works~\cite{hou2022self,mehta2023privacy}. The transformer-based pre-training approach can learn robust and generalizable features, taking advantage of large unannotated data from medical databases of past seizure records.
To ensure generalization when annotated examples are scarce, the module can be flexible by incorporating emerging approaches such as meta-learning~\cite{mahajan2020meta} and domain adaptation~\cite{dissanayake2020domain} which are becoming particularly valuable in the medical domain.

\vspace{-6pt}
\subsection{Multi-stream frameworks for seizure classification} 
After the presence of an epileptic seizure is confirmed by the preceding module, the focus of the ``Process'' module shifts to the analysis of diverse clinical manifestations, including isolated semiologies captured from one or more cameras. The models employed in this module are designed to serve as detectors for common lateralization and localization signs, including phenomena such as head version, mouth automatism, dystonic limb posturing, hand automatisms~\cite{ataouglu2015evaluation,ferando2019hand,stefan2023ictal}, and even specific signs like the thumbs-up gesture~\cite{vilaseca2021thumb}.
The multi-stream framework exhibits flexibility in capturing dynamic changes through appearance, keypoints, and the analysis of the 3D distribution of body parts, face and hand landmarks, employing either a fusion approach~\cite{ahmedt2018deep,hou2022automated} or a hierarchical approach~\cite{ahmedt2018hierarchical}. 
Depending on the specific downstream task, such as classifying GTCS, MTLE, ETLE, etc, users have the ability to identify the most relevant motion features deserving attention for a given patient, allowing customization of the analysis for the complete body~\cite{karacsony2022novel,hou2022self}, face~\cite{ahmedt2019vision,hou2022automated}, or hands~\cite{pemasiri2019semantic}. 
Recognizing that a single semiologic sign in isolation will likely not provide sufficient insight, the integration of validation accuracies for each clinical manifestation is calculated. This consolidated information supports clinical experts in making the final decision. Furthermore, the observed ictal behavior can be correlated from a semi-automated perspective with neuronal activity findings, whose concordance is necessary for localization purposes, as has been demonstrated by preliminary research in~\cite{hou2020rhythmic,zalta2020neural}. 
Methodologies discussed in this module can be further enhanced with open-set action recognition strategies~\cite{BaoICCV2021DEAR} that require the network to correctly classify in-distribution samples and identify out-of-distribution samples. This is discussed further in Section~\ref{opp_open-set}.

\vspace{-6pt}
\subsection{Model prediction analysis and interpretation} 
This module serves a dual purpose. Firstly, it aims to provide insights into the confidence levels associated with each classification, addressing the potential perception of deep models as a ``black box'' by conveying the uncertainty inherent in a model's decisions. Secondly, it strives to offer model-agnostic interpretations, facilitating a deeper understanding of predictions by providing contextual information.
Visualizations are employed to showcase the variety of semiological patterns that are present in the database, highlighting dissimilarities between patterns exhibited by different patients. This visual representation reflects the classifier’s confidence and aids in the interpretation of ``correct'' or ``incorrect'' classification decisions~\cite{ahmedt2018hierarchical}. The incorporation of Bayesian deep learning~\cite{kendall2017uncertainties} or deep ensemble approaches~\cite{lakshminarayanan2017simple} in this module is integral to the uncertainty quantification process.

Recent works in vision-based seizure semiology (Table~\ref{table:semiology}) and review papers~\cite{karacsony2023deep,knight2023artificial} confirm the current lack of explainability within existing methods. This absence of ``transparency'' poses a significant barrier to the complete adoption of AI in clinical practice, as physicians are hesitant to trust machine learning model predictions without sufficient evidence and interpretation, especially in disease classification. This module incorporates attribution-based methods, popular in the medical domain for their model-agnostic plug-and-play nature. To support module development, studies involving expert clinicians to rate explanations and assess the limitations of model-agnostic interpretation methods are crucial~\cite{arbabshirani2018advanced}. Further studies should explore clinical workflow integration and investigate how clinical experts could refine model decisions, as demonstrated with the Nelli\textregistered hybrid system~\cite{basnyat2022clinical} for seizure detection. Additional research strategies in interpretability for human action recognition are outlined in Section~\ref{opp_explain}.

\vspace{-6pt}
\subsection{Anomaly detection - aberrant epileptic seizure} 
Recognizing actions (motor behaviors) in the context of epileptic seizures is challenging due to their subjective and dynamic nature. Different clinical manifestations, such as oral and manual automatism, can have a large amount of variation in their execution, and may sometimes be related to normal behavior, making it difficult for recognition models to generalize effectively. 
While a larger labeled or unlabeled semiology database could potentially enhance accuracy, especially in situations where technology and clinical expertise are available, we aim to augment our system with a ``Feedback'' module. 
This module is designed to detect unusual or aberrant behaviors, contributing to the overall robustness and adaptability of the overall framework. 
In addressing this task, epileptic seizures are grouped into a best-fit model using a template of stereotypical behaviors commonly observed in various forms of epileptic seizures, organized into libraries. This allows the system to identify whether the test semiology can fit into within a set of known categories~\cite{ahmedt2019aberrant}.  Semi-supervised video anomaly detection techniques, such as deep autoencoders, also prove effective in this context, leveraging the full potential of normal data while addressing issues of weak labeling and data imbalance~\cite{nayak2021comprehensive}.  However, the selection of the most suitable deep-learning method for video anomaly detection depends on the available datasets.

Through continuous monitoring and identification of deviations from typical epileptic behaviors, the system dynamically evolves using an active learning mechanism. This iterative process enables the system to update its understanding of abnormal patterns, improving its ability to classify and interpret diverse manifestations. When the module detects an aberrant behavior, a human-in-the-loop review process involving clinical experts~\cite{tian2020graph} assesses the nature of the abnormality. The human-reviewed instances are integrated back into the training dataset, facilitating the deployment of a refined model.

\vspace{-6pt}
\subsection{Future implementation}
In the future, we can imagine such a system being incorporated into the Epilepsy Monitoring Unit video set-up such that clinicians can both review the video-EEG data in the manner they currently do, and benefit from a user-friendly interface that would allow additional objective automated video analysis to complement their own visual assessment (similar to the current clinical use of automated EEG or MRI quantification software). Such a system could, for example, automatically measure the frequency of seizure-related rhythmic movements, compare the similarity index between different seizures from the same patient, compute a predictive score for diagnosis (\textit{e.g.}~ES versus PNES), or predict likely cerebral localization and/or lateralization with estimation of confidence intervals. This would be most useful for complex motor semiology (\textit{e.g.}~patterns seen in hyperkinetic seizures)~\cite{fayerstein2020quantitative}.  

Self-learning systems based on deep learning approaches will become more accurate as data accumulates, progressively allowing for more precise categorization of different semiologic patterns, including complex ones that are currently challenging for clinicians to interpret.  As privacy-preservation methods become more advanced and methods start to be used on a larger scale with more collaboration and data sharing between clinical teams, knowledge of neural correlates of different semiologic patterns will expand. Knowledge sharing will likely be helped by similar quantified methods being applied to simultaneous electroencephalographic data (especially stereo-electroencephalography, SEEG), overseen by expert clinicians and using seizure freedom after epilepsy surgery as an additional ground truth where appropriate. 
Once validated in specialist centers, the developed techniques may then be applicable to increasingly apply automated video analysis of seizures in ambulatory settings as well, maximizing access to accurate seizure diagnosis and meeting a critical current gap in Epilepsy care. Beyond the Epilepsy monitoring unit environment, other seizure video recording methods including smartphones~\cite{zuberi2021multi} might benefit from some inbuilt automated tools, with lightweight processing capacity, to complement the clinicians' visual analysis, which could make a precise and timely diagnosis of seizures more widely available.



\section{Research opportunities for video-based seizure analysis}
\label{opport}

\subsection{Clinical challenges of motion capture and domain adaptation}
\label{opp_mocap}

Motion analysis is one of the main approaches that can assist medical practitioners in analyzing semiology. In~\cite{karacsony2023deep}, authors reviewed MoCap methods in relation to medical challenges including occlusions, viewpoints variations, and low-resolution 3D MoCap; and suggested paradigms that can be considered to address these challenges. Here, we provide some outline MoCap challenges and guidelines that can be leveraged to assist with semiology analysis. The repertoire of movement patterns and facial expressions observed in human seizures is extremely large, complex and heterogeneous compared to those studied in animal models~\cite{knight2023artificial}, thus  ideally suited to benefit from machine learning approaches but also very challenging in terms of obtaining adequate test data sets with expert clinician labelling.

The MoCap information obtained from face detection and tracking and facial landmarks can be beneficial for semiology analysis. In Table~\ref{table:CV_techniques}, works that have applied computer vision algorithms for facial MoCap are listed. In the majority of these works, the authors considered 2D approaches for their analysis, whereas more information could be inferred from 3D facial analysis. For example, Ahmedt-Aristizabal et al.~\cite{ahmedt2019vision} leveraged a 2D to 3D computer vision algorithm~\cite{feng2018joint} to include rich facial (mouth) information in their analysis. 
Advances in computer vision algorithms facilitate the estimation of 3D locations from 2D videos/images (\textit{e.g.}~\cite{sharma20223d} reviews SOTA models for 3D face reconstruction using deep learning). These new approaches allow high-quality 3D face reconstructions, which in return provides more information such as dynamic facial changes that can be used in semiology analysis.  

The hand MoCap pipeline used in previous semiology studies~\cite{ahmedt2018hierarchical,ahmedt2019motion} has demonstrated promising performance, though the most recent computer vision methods are yet to be investigated for semiology videos. For example, in recent works domain adaptation along with self-supervision has been used to assist with autism spectrum disorder (ASD) diagnosis~\cite{yoo2023pointing}, which is a challenging task in medical AI. Other examples of more advanced hand MoCap are analyzing 3D poses~\cite{cheng2023handr2n2,fan2020adaptive}, or methods that deal with hand occlusion~\cite{zhang2023ochid}.

In addition to the aforementioned recent techniques for face and hand MoCap, increasingly advanced algorithms for human pose extraction can be explored on semiology data. 2D human pose estimation can be inferred through transformers (ViTPose)~\cite{xu2022vitpose}, while 3D pose estimation methods have been proposed using graph neural networks (GCN)~\cite{zhao2022graformer}. Methods proposed by~\cite{zheng20213d} and ~\cite{zhou2023diff3dhpe} can leverage spatial and temporal information from sequences of semiology videos to improve pose estimation. 
Test-time adaptation~\cite{lee2023tta}, domain adaptation~\cite{bigalke2023anatomy} or knowledge distillation~\cite{mehta2023privacy,hou2022automated} are helpful when there is no access to the source data for training, or in cases where only a small dataset is available.

Advancements in computer vision MoCap methods have demonstrated high-performance gains, particularly on public datasets. For example, in domains like 3D face reconstruction and 3D pose estimation, novel techniques such as transformers or graph-based methods have been introduced. These methods show promise in addressing challenges such as occlusion and providing a more robust representation of dynamic features in faces, hands, or the body. The advancements, which have shown success in various applications, warrant exploration in the context of seizure semiology analysis.
As experience and sensitivity/specificity of methods improved, more subtle discrimination will be possible, \textit{e.g.} distinguishing between ASD-related hand stereotypies and epileptic movements in a patient with both conditions. As such, future work may concern the prediction of depression scores from facial expression analysis etc, in patients with epilepsy, i.e. detecting presence of comorbidities in the resting state (between-seizure periods), which could be considered as ``inter-ictal semiology''.

\vspace{-6pt}
\subsection{Preserving privacy in patient monitoring for Epilepsy}
\label{opp_privacy}

The application of AI in healthcare presents ethical dilemmas that require careful consideration. A primary ethical concern for healthcare AI is privacy and data security. As AI algorithms depend on extensive patient data for training and optimization, securing sensitive patient data and limiting access to authorized personnel only is essential. Recent reviews concerning AI and medical AI privacy-preservation~\cite{khalid2023privacy,ravi2023review} demonstrate different techniques that can be used to protect the privacy of patients' data. Here, we briefly explain visual approaches known as obfuscation that can be used to protect semiology videos. 

Visual techniques, also known as de-identification techniques, are designed to enhance visual privacy. The prevalent approach in semiology data involves transforming the image or video data into another modality, such as converting video to optical flow. These pipelines~\cite{geertsema2018automated,van2020automated,garccao2023novel,mehta2023privacy} can operate locally within the hospital, safeguarding patient privacy while providing motion semiotics of seizures. However, they fall short in providing context for the analysis of specific behaviors of interest, such as small facial or hand semiology. For example, facial seizure manifestations (\textit{i.e.}~involuntary seizure-induced gaze deviation, specific facial contraction patterns, or emotional facial expression) may bring key information to clinical seizure analysis are often not visible in optical flow images.

Another visual-based approach is image swapping/synthesis, which is employed to conceal a patient's identity while conveying information not easily leveraged by optical flow-based approaches. For instance, ~\cite{zhu2020deepfakes} developed a face-swapping method for patients with Parkinson disease, and demonstrated that face swapping could preserve the facial landmarks while hiding the patient's identity.
The efficacy of different face transformation approaches was then investigated in patients with facial semiology captured during epileptic seizures~\cite{hou2023artificial}. Swapping the whole body with an avatar can also be beneficial for privacy protection, and such techniques are based on estimating landmarks/pose, generating a high-quality mesh, and replacing the human body in a video or image with an avatar~\cite{rong2021frankmocap,pavlakos2019expressive}. Image or video synthesis can also be used at a higher level to generate synthetic data for semiology. For instance, advanced methods that take text~\cite{tevet2022human} or physics-based approaches~\cite{yuan2023physdiff} along with semiology expert knowledge can be investigated to generate such data.   

Image distortion, such as blurring or pixelation, is an image-level technique that can be leveraged to preserve patients' identities. Blurring or pixelation involves applying a filter to a specific part of an image, such as the face to make the subject unidentifiable. While these alterations can partially preserve privacy, they are vulnerable to reconstruction attacks~\cite{mai2018reconstruction} and are deficient in preserving MoCap semiology features~\cite{hukkelaas2023does}.    

Privacy preservation can also be achieved at the model or system level through design to render patients unidentifiable. In this approach, data, algorithms, or both are structured to prevent mechanisms relating to facial analysis from revealing the patient's identity, while maintaining high performance of the action recognition pipeline. For instance, \cite{ren2018learning} developed an adversarial training paradigm to anonymize human faces in videos without affecting action recognition performance. Another example involves leveraging homomorphic encryption~\cite{kim2022secure}, which retains action detection performance while ensuring information confidentiality. 

Federated learning has gained widespread attention for its capacity to train machine learning models while preserving privacy. Positioned as a system-level privacy mechanism, federated learning ensures the protection of patients' privacy by keeping their information localized, eliminating the need for data interchange. This decentralized approach is designed to learn from heterogeneous datasets, converging into a global model that upholds local privacy standards~\cite{khan2021federated,khokhar2022review}. Li et al.~\cite{li2021meta} proposed a federated learning approach to train action recognition models without central collection of users' sensor data.  

Finally, visual privacy can also be achieved at the sensor level. For example, extremely low-resolution cameras have been explored for action recognition~\cite{dai2015towards,ryoo2018extreme,bai2023extreme,wang2023modeling}. Videos obtained from these low-resolution cameras can preserve patient privacy while enabling the inference of semiological actions. 

The application of privacy-preserving paradigms in semiology is not only beneficial for patients but also for the computer vision community, allowing data and models to be shared and benchmarked to improve the performance of current algorithms. In addition, a memory bank or profile can be built for each patient, which can be shared among semiology experts for further analysis.

\vspace{-6pt}
\subsection{Fine-grained analysis of actions for the propagation of semiology}
\label{opp_fine-grained}

Epileptic seizures are characterized by multiple clinical features which transition over a short timescale (seconds-minutes), reflecting the anatomical spread of the electrical seizure activity, in both its spatial (anatomic) and temporal (\textit{e.g.}~discharge frequency) aspects~\cite{mcgonigal2021seizure}. Knowledge of clinically plausible seizure types and propagation pathways can therefore make a valuable contribution to the differentiation of focal seizure types, their differential diagnoses and the cerebral organization of seizure activity~\cite{noachtar2009semiology}.
Existing action recognition models, as discussed in Section~\ref{litreview}, often focus on recognizing full video segments without capturing subtle transitions or dependencies between different clinical manifestations. This limitation hinders the understanding of the temporal evolution of seizures from onset to termination.

Human action recognition offers an avenue to capture the evolution of semiology, and such approaches fall into two categories: discrete methods that operate on pre-segmented videos (current research in vision-based semiology); and continuous fine-grained methods consider not only the actions but also capture the transitions between them. While discrete methods demonstrate higher performance, they are disconnected from real-world scenarios composed of fine-grained actions. Fine-grained action analysis allows tracking of previously observed actions and exploring relationships between consecutive actions, capturing both inter (long term) and intra (short term) sequence relationships~\cite{gammulle2023continuous,ding2023temporal}.

In~\cite{lea2017temporal} a temporal convolutional network (TCN) segments a sequence of human actions into action segments, capturing information such as action durations, pairwise transitions, and long-range action dependencies in the actions performed. The multi-stage TCN model (MS-TCN)~\cite{farha2019ms} enhances the temporal resolution by aggregating information across multiple-stages. A further extension, boundary-aware cascade networks, addresses issues related to inaccurate action segmentation boundaries and misclassified short action segments~\cite{wang2020boundary}. However, these networks rely on detailed information about action boundaries which may be impractical for seizure analysis due to the dynamic nature of seizures (\textit{i.e.}~for human action segmentation each action is temporally localized with precise start and end times). Action segmentation models for seizures must minimize reliance on dense frame-level supervision due to the difficulties in defining the start and end of specific semiologies. Towards this, some preliminary semi-supervised and pseudo-labeling strategies were introduced in~\cite{gammulle2020fine,singh2021semi}.

In the context of semiology analysis, weakly supervised action segmentation, which utilizes signals like ordered and unordered action lists without frame-by-frame labels, becomes more relevant.
To exploit a sequential list of actions, some works adjust action boundaries gradually with a soft labeling scheme~\cite{ding2018weakly}, or train class-specific discriminative action prototypes~\cite{chang2021learning}.
Considering weaker forms of supervision which lack action ordering and frequency information, \textit{i.e.}~how many repetitions of each action occur~\cite{richard2018action}, authors have learned a segmentation network directly from the action sets using a set prediction loss~\cite{fayyaz2020sct}, or introduced a pairwise order consistency loss to penalize disagreements in ordering between the extracted templates and the segmentation model outputs~\cite{lu2022set}.

Action segmentation is a rapidly evolving topic that holds promise in representing the evolution of semiology, and offers several directions that will drive progress and innovation in the field.

\vspace{-6pt}
\subsection{Activity recognition in open-set environments}
\label{opp_open-set}

Action recognition models, despite being valuable and effective in numerous medical applications, have a number of drawbacks. These limitations include the incapability to generalize to unfamiliar actions, an inability to accommodate new action categories, and difficulties in handling ambiguous actions. Consequently, these limitations have spurred the emergence of open-set action analysis approaches. These approaches aim to address these challenges by recognizing both known and unknown actions, thereby enhancing robustness and establishing more scalable and reliable recognition systems. 

DEAR~\cite{BaoICCV2021DEAR}, an open-set action recognition model proposed by Bao et al., addresses these issues by estimating the uncertainty associated with individual labeled actions. This estimation of uncertainty is instrumental in differentiating between known and unknown samples. Consequently, the deep neural network recognizes and quantifies uncertainty in its predictions.
MULE~\cite{zhao2023open}, a multi-label evidential learning approach for open-set action recognition, leverages the concept of evidential learning. Specifically, evidential learning allows for the quantification of uncertainty in predictions. By using a multi-label formulation, the model is capable of assigning multiple labels to an input, including both known and unknown actions.
Recently, ~\cite{jun2023enlarge} examined the behavior of feature representations in the context of open-set action recognition. Drawing inspiration from information bottleneck theory, ~\cite{jun2023enlarge} proposed a method to enhance the performance by enlarging the instance-specific and class-specific information present in a feature. The enhancement component for class-specific information leverages a classifier that has been trained to refine feature representations, thereby increasing the discriminative power of the model. Through this approach, the model explicitly models the distinctions between known and unknown classes, leading to more effective differentiation during the recognition process.

The exploration of open-set action analysis has yielded promising strategies to overcome limitations in traditional action recognition models. These advancements signify a meaningful stride forward in the continuous evolution of seizure semiology analysis.

\vspace{-6pt}
\subsection{Explainable patient activity recognition based on interpretable models}
\label{opp_explain}

Machine learning models developed through deep learning techniques are often regarded as complex and difficult to interpret, typically necessitating specialized methods for understanding their inner workings. Gaining an understanding of deep learning models, often regarded as enigmatic ``black boxes'', can be achieved through techniques such as analyzing feature importance~\cite{lundberg_nips2017}, visualizing activations~\cite{alex_nap_2022}, employing Grad-CAM~\cite{selvaraju2017grad}, utilizing saliency maps~\cite{kadir2001saliency}, guided backpropagation~\cite{springenberg2014striving}, applying Local Interpretable Model-agnostic Explanations (LIME)~\cite{ribeiro2016should,tomi_lime_2018}, utilizing t-SNE for data visualization~\cite{tsne_2008}, optimizing activation patterns~\cite{tan2023visualizing}, conducting model-specific visualization~\cite{Chefer_2021_CVPR}, and incorporating attention mechanisms~\cite{guo2022attention}. These techniques provide insights into the decision-making processes of deep learning models, with a particular emphasis on spatial data inputs. This contributes to enhancing model transparency and reliability across diverse applications. However, these methods are most applicable to spatio-temporal data to gain spatial insights. 

One of the major hindrances in generating spatio-temporal interpretation outputs is the lack of a model-agnostic interpretation pipeline. This is challenging, as the spatio-temporal interpretation illustrates the significance of detecting biases in video data and action associations through deeper network analysis. 
The motion-guided sampler (MGSampler)~\cite{zhi2021mgsampler} attempted to interpret the dynamic motion characteristics by leveraging motion as a fundamental signal for frame selection. It utilizes a frame selection approach that ensures an even distribution of frames across important segments, characterized by high motion salience, achieved through the accumulation of motion data.
M-MiT~\cite{monfort2021multi} provides a more comprehensive method that is focused on interpreting action recognition models. This method is focused on examining these interpretable features and how they contribute to a network's output. It reveals insights that might otherwise remain concealed without the capacity to identify underlying action concepts.

Graph-based models offer an intuitive representation that is easily interpretable. However, learning directly from the data can pose challenges for graph-based models. A recent approach~\cite{zhong2022spatiotemporal} utilizes a graph structure to represent the interactions between different body parts throughout the motion sequence. The graph construction is based on the spatial relationships among body joints, enabling the model to capture the influence joints have on one another's movements. This spatial information facilitates the interpretability of the model in terms of spatial relationships. Furthermore, the sequential nature of the motion data allows for the modeling of temporal relationships, enhancing the model's interpretability in terms of temporal aspects.

Predicting human motion in future frames is another widely used approach to interpret dynamic sequences and their corresponding actions. This technique allows us to gain valuable insights into the model's training focus and potential biases towards the specific data it has been trained on. ~\cite{tang2023m2ast} have successfully achieved an adaptive and concise representation for diverse graph structures of human motion when modeling the spatio-temporal actions using complex structured graphs. This graph-based approach incorporates the spatial and temporal dependencies by considering the sequential nature of the motion data which allows the model to effectively capture the temporal relationships between different frames in the motion sequence. Additionally, this framework can establish temporal dependencies in the subsequent frames, which can contribute to model interpretability.

A natural question that arises is if the decision-making process in deep learning models for action recognition in semiology can be interpretable. Considering the spread of deep processing for various medical applications, explainability and quantitative evaluation with a focus on usability by clinicians is crucial~\cite{kundu2021ai}. A next step is to combine seizure video automated analyses with other investigations including anatomo-electroclinical correlations from stereoelectroencephalography (SEEG) data~\cite{bartolomei2017defining}, which could bring new insights via multi-modal artificial intelligence approaches~\cite{karimi2023generalisability}. Once sufficiently large and diverse human datasets are compiled for testing, such analyses could ultimately refine the power of automatically detected seizure video patterns to correctly classify seizure type and predict likely cerebral localisation.

\section{Conclusion}

Recent advances leveraging deep learning models for the detection and classification of seizure semiology show great promise. This survey provides insights from existing works and outlines several challenges and opportunities that will benefit the field and other medical applications. The proposed pipeline introduces a comprehensive integrated system, aiming to quantify and appropriately evaluate video recordings of seizures. 
As the field progresses, the exploration of multi-sensory methodologies capable of learning across semiology, electrical patterns, and neuroimaging for precise localization emerges as a key challenge for the neuro-engineering scientific community. 
While clinical expertise at the individual and team levels remains irreplaceable, the provision of more quantitative, complementary evidence holds the potential to enhance medical decision-making and ultimately elevate the standard of care for patients.

\bibliographystyle{splncs04}

{\scriptsize
\bibliography{ref}}




\end{document}